\documentclass[11pt]{article}

\usepackage[utf8]{inputenc}
\usepackage[T1]{fontenc}

\usepackage{arxiv}

\usepackage{amsfonts}
\usepackage{amsmath}
\usepackage{amssymb}
\usepackage{nicefrac}
\usepackage{booktabs}
\usepackage{multirow}
\usepackage{graphicx}
\usepackage{subcaption}
\usepackage{enumitem}
\usepackage{natbib}

\usepackage{hyperref}
\usepackage{url}
\hypersetup{
  colorlinks=true,
  linkcolor=arxivaccent,
  citecolor=arxivaccent,
  urlcolor=arxivaccent,
  filecolor=arxivaccent,
  breaklinks=true,
}

\title{User as Engram: Internalizing Per-User Memory as Local Parametric Edits}

\author{%
	Bojie Li \\
	Pine AI
}

\date{}
\runningtitle{User as Engram: Internalizing Per-User Memory as Local Parametric Edits}

\begin{document}

\maketitle

\begin{abstract}
Personal memory in a language model is two problems, not one:
\emph{content} (the specific facts about a user) and
\emph{reasoning skill} (the ability to turn those facts into
answers). The brain keeps the two apart (a sparse, local
\emph{engram} in the hippocampus for each episode, a slow neocortex
for the shared skills that interpret it), so a new fact need not
overwrite everything else. Most personalization today keeps a
user's facts \emph{outside} the weights, in a natural-language
memory file or a retrieval index. When facts are written
\emph{into} the model instead, the standard recipe is the per-user
LoRA adapter, which does the opposite of the brain, folding
content and skill into one global weight delta.
Writing a user's facts as a LoRA contaminates text unrelated to them;
writing the same facts as local \emph{Engram rows} leaves it
mathematically untouched, resulting in a roughly $33{,}000\times$ smaller
memory footprint.

We therefore propose \textbf{User as Engram}: store a user's
content as surgical edits to the hash-keyed memory table of an
Engram model, and carry the reasoning skill in one \emph{shared}
adapter. This layered
design matches per-user LoRA's direct recall while delivering
$5.6\times$ higher indirect-reasoning accuracy on average,
and never makes a single user worse at reasoning than
the untouched base. The edit is a glass box: writing a fact switches
on its lookup at exactly the trigger, adds the value the answer needs,
leaves every other position unchanged to the last bit, and fails if
written into the wrong layer.
Because different users' facts land in disjoint hash slots, their edits
\emph{compose}: many users live in one shared table at once, stacking
additively and losslessly, where a per-user LoRA, a single global
weight delta, admits only one.
Upon retrieval, a per-user Engram table does not grow with the
population the retriever must search, so past $\sim$100 facts it overtakes a retrieval pipeline on a
$2.5\times$ larger model.
\end{abstract}

\begin{center}
\small
Code: \url{https://github.com/19PINE-AI/user-as-engram} \\[2pt]
Website: \url{https://01.me/research/user-as-engram}
\end{center}
\vspace{-0.4em}

\section{Introduction}
\label{sec:intro}

Maya has talked to her AI assistant for the better part of a
year. Over those months it has learned that her cardiologist is
Dr.~Elena Vasquez, that Vasquez practices at Lakeside
Cardiology, that Maya is severely allergic to penicillin, and
that she went vegetarian in 2024. A useful assistant must do
three things with this \emph{personal memory}: recall a fact on
demand (\emph{``who is my cardiologist?''}), reason over it when
the question is indirect (\emph{``I'm visiting my daughter in
California next month, where should I go for a heart
check-up?''}), and never let one user's facts leak into
another's. Serving millions of such users from a single
transformer~\citep{vaswani2017transformer} backbone is an open
architectural problem.

Human memory already solves a version of it. The brain does not
record a new fact by nudging every synapse a little; it writes
the episode as a sparse, local trace (an \emph{engram}) in the
hippocampus, while the slow, distributed neocortex supplies the
general skills that interpret
it~\citep{semon1921mneme,tonegawa2015engram,mcclelland1995cls}.
Keeping the fast, local store apart from the slow, shared one is
exactly what lets us learn that Maya is allergic to penicillin
without disturbing how we reason about allergies in general.

\begin{figure}[t]
  \centering
  \includegraphics[width=\linewidth]{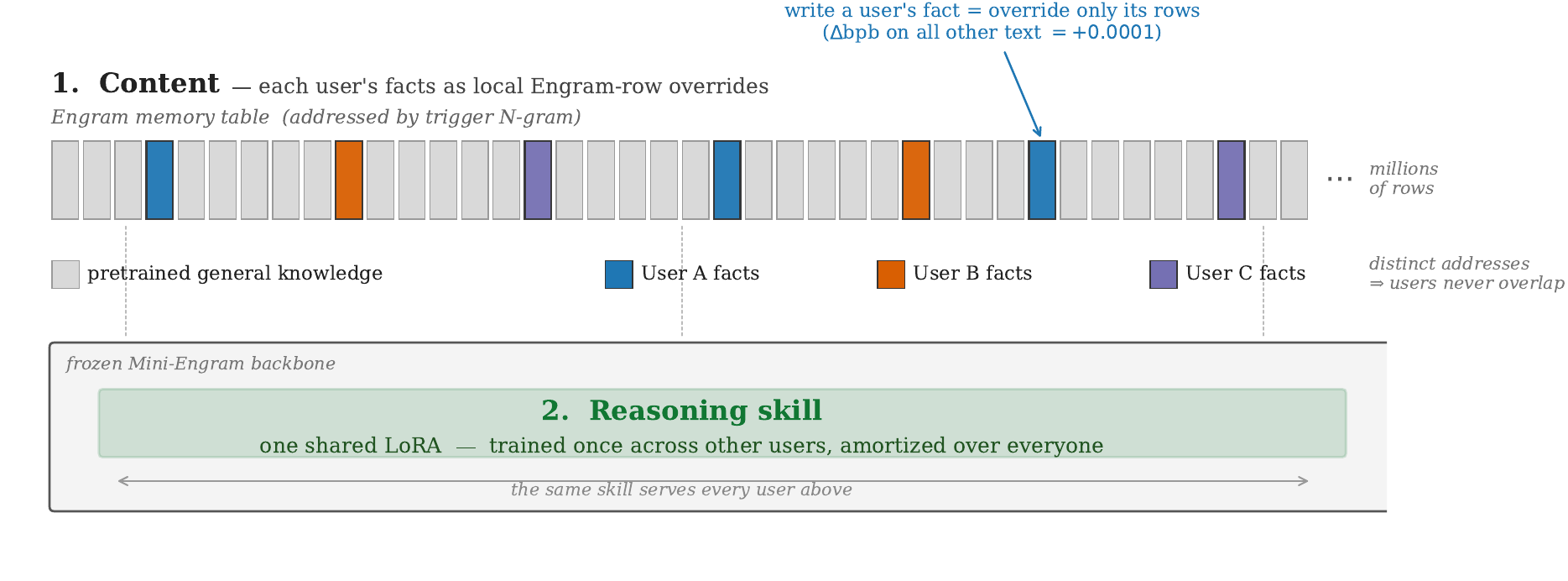}
  \captionsetup{skip=4pt}
  \caption{\textbf{User as Engram} splits personal memory into two
  layers, mirroring the brain's complementary learning systems.
  \textbf{Top (content).} Each user's facts are written as a few local
  row overrides (colored) in the Engram model's one hash-keyed memory
  table; the table's other rows hold the general knowledge learned at
  pretraining (gray). A write touches only that user's rows
  ($\Delta$bpb${}=+0.0001$ on all other text), and different users hash
  to different addresses, so they never overlap: the fast, sparse,
  local trace of a hippocampal engram.
  \textbf{Bottom (reasoning skill).} A single shared LoRA on the frozen
  Mini-Engram backbone, trained once across other users and amortized
  over everyone: the slow, shared neocortex. The only per-user state is
  the handful of colored rows; everything below is shared.}
  \label{fig:layered-arch}
\end{figure}

Personal memory in a language model has the same two-part structure,
\emph{content} (the user's specific facts) and \emph{reasoning skill}
(turning facts into answers), and the two pull in opposite directions.
Content is private and differs from user to user, so it wants its own
local store; the reasoning skill is common to everyone, so it should
be learned once and shared (Figure~\ref{fig:layered-arch}).

Three families of methods address personal memory. In deployed
systems the two that dominate are both non-parametric: in-context
learning (ICL,~\citealt{brown2020gpt3,min2022rethinking}), most often
an automatically-extracted natural-language memory file, keeps the
facts in the prompt, and
retrieval-augmented generation
(RAG,~\citealt{lewis2020rag,gao2023ragsurvey}) fetches them at
query time. Both leave the model's weights untouched, and we
compare against retrieval head-to-head throughout
(Sections~\ref{sec:layered-rag}--\ref{sec:rag-scale}). The third
family writes the facts \emph{into} the weights, and only this one
conflates content and reasoning skill, storing both in the
\emph{same} parameters.

Among in-weights methods the standard recipe is a per-user
LoRA~\citep{hu2022lora} trained on each user's facts
(Section~\ref{sec:background}). A LoRA has no address: to
make one fact more likely, gradient descent bends the $Q/K/V/O$
directions that fit it most cheaply, and those same directions fire on
unrelated text, so the edit is global by construction, not a
side-effect of training. A global delta is also a single-tenant one:
two users' LoRAs cannot share a model without merging into a combined
delta that interferes, so each user needs their own copy of the
weights.

\paragraph{Our approach: content in a memory table, skill in a shared adapter.}
\textbf{User as Engram} writes each of Maya's facts as a surgical edit
to the hash-keyed memory table of an Engram-pretrained
model~\citep{cheng2026engram}, whose gated lookups fire only on suffix
N-grams via deterministic hashing; the reasoning skill is one
\emph{shared} LoRA trained once across a held-out population, so the
only per-user state is the memory rows, swapped in per request. The
table is \emph{content-addressable} and inert off its key, so writing a fact
changes the output only where its trigger N-gram is read
(Section~\ref{sec:locality}, Figure~\ref{fig:locality}). That is why
the table, not the weights, is the right home for a user's facts:
writes are isolated, the
suffix-N-gram index matches how personal facts are queried (by their
surface form), and each row is a fixed $\sim$88\,KB whose address swaps
in per request, so storage grows with users, not model size.
Isolation buys composability for free: because different users occupy
disjoint addresses, their override maps commute and stack additively
into one shared table, putting many users (or a corporate store layered
under a personal one) in the same model at once (Figure~\ref{fig:layered-arch},
Section~\ref{sec:additive}).

\paragraph{Organization of the paper.}
The rest of the paper develops four core claims:
\begin{enumerate}[leftmargin=*,topsep=2pt,itemsep=2pt]
  \item \textbf{The cost of storing facts in shared weights}
        (Section~\ref{sec:negative}).
        On a controlled base, writing a fact as a LoRA rather than as
        an Engram row is a $\sim$33{,}000$\times$ difference in extra
        loss on unrelated text (3-seed mean), a property of how a
        LoRA stores a fact, not of how we tuned it. The extra loss is
        always there; the visible damage (worse reasoning) appears
        only on weak bases, and vanishes on a strong instruction-tuned
        base that answers the question on its own. We turn that damage
        on and off by changing base strength.
  \item \textbf{User as Engram, with the mechanism opened up}
        (Section~\ref{sec:method}):
        a way to write a user's facts as small, local edits to an
        Engram model's memory table. Unlike a LoRA, whose change is
        spread invisibly across the whole model, every step of an
        Engram write can be watched on the trained model: it switches
        on the memory lookup at exactly its trigger, adds the value
        the answer needs and nothing else (every other position is
        unchanged to the last bit), and stops working if written into
        the wrong layer (Figures~\ref{fig:locality} and~\ref{fig:glassbox}).
        We release four Mini-Engram checkpoints
        (178\,M--1.22\,B; a nano-scale public replication of
        Engram), all code and data, and a 50-line multi-tenant
        server with zero cross-user leakage.
  \item \textbf{A layered design} (Section~\ref{sec:layered})
        that pairs per-user Engram content with one shared
        reasoning adapter (an artificial complementary learning
        system) and beats every all-in-one baseline on direct
        recall, indirect reasoning ($5.6\times$ over per-user
        LoRA on average), contamination, and per-user reasoning regressions.
        The lead survives a personal $\to$ medical schema shift.
  \item \textbf{A deployment-scale comparison with retrieval}
        (Sections~\ref{sec:layered-rag}--\ref{sec:rag-scale}):
        which method wins is set by facts-per-user and population
        size, not benchmark size. Because a per-user Engram table
        does not grow with the population, it overtakes a strong
        retrieval pipeline (running on a $2.5\times$ larger
        model) once the knowledge base passes $\sim$100 facts.
\end{enumerate}

\section{Background}
\label{sec:background}

\paragraph{The Engram architecture.} \citet{cheng2026engram}, part
of the DeepSeek line of sparse
models~\citep{dai2024deepseekmoe,deepseekai2024v3}, introduce
\emph{conditional memory} as a form of sparsity that complements
mixture-of-experts (MoE,~\citealt{shazeer2017moe,dai2024deepseekmoe}).
At each token position $t$, suffix N-grams of canonicalized tokens are
hashed via $K$ multiplicative-XOR heads into prime-sized embedding
tables $E_{n,k}$. Retrieved rows $e_t$ are projected by learned
$W_K, W_V$ and gated by an attention-style scalar
$\alpha_t = \sigma(\mathrm{RMS}(h_t)^\top \mathrm{RMS}(W_K e_t)/\sqrt{d})$.
The output $\alpha_t W_V e_t$ is added to the residual stream at the
selected Engram layer. Crucially, the address $z_{t,n,k}$ is
deterministic from the input token IDs---known before the forward
pass---so the table can be offloaded to host DRAM without GPU
contention.

\paragraph{LoRA-as-memory family.} Per-user/per-document LoRA
adapters~\citep{su2024prag,tan2024oppu,zhuang2024hydra,t2l2024,
tan2024perpcs,memlora2025} train a low-rank
delta~\citep{hu2022lora,houlsby2019adapters,mangrulkar2022peft}---one
member of a broad parameter-efficient fine-tuning family that also
includes prefix-tuning~\citep{li2021prefix} and adaptive-rank and
quantized variants~\citep{zhang2023adalora,dettmers2023qlora}---on
$Q/K/V$ (and sometimes MLP) matrices on top of a frozen base.
It is fit per user via NTP on a
(observation, fact, QA) mixture,
via knowledge distillation from a teacher
(DyPRAG~\citep{tan2025dyprag},
DistilledPRAG~\citep{distilledprag2025}), or via hypernetwork
emission (T2L~\citep{t2l2024}). All variants \emph{edit the model
weights globally}: every forward pass sees the LoRA's change.

\begin{figure}[t]
  \centering
  \includegraphics[width=0.82\linewidth]{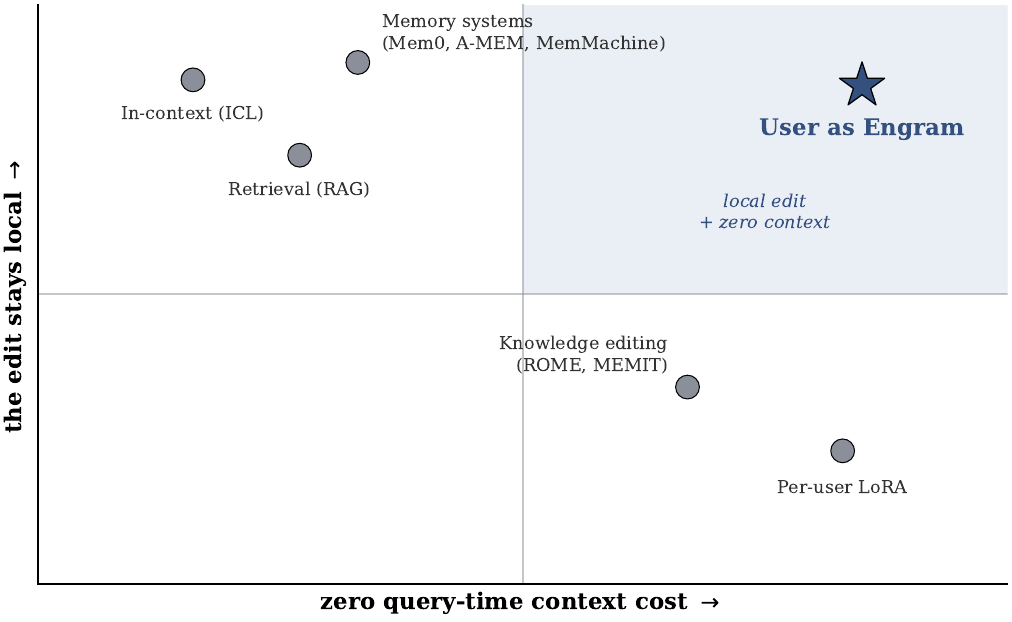}
  \caption{Where User as Engram sits among personal-memory methods.
  Context-based methods (in-context learning, retrieval, memory
  systems) leave the model untouched but pay context tokens at query
  time (top-left); weight-based methods (per-user LoRA, knowledge
  editing) pay no context but edit the model globally (bottom-right).
  User as Engram occupies the remaining corner: a local edit at zero
  context cost. Axes are qualitative.}
  \label{fig:landscape}
\end{figure}

\paragraph{Where User as Engram sits.}
These two---the Engram store and the per-user LoRA---anchor the
landscape of personal-memory methods, which sorts along two axes
(Figure~\ref{fig:landscape}): whether a method pays context tokens at
query time, and whether storing a fact leaves the rest of the model
untouched. In-context learning, retrieval, and external memory systems
keep the model local but pay context on every query; per-user LoRA and
knowledge editing pay no context but change the weights globally. User
as Engram is the only one in the remaining corner---a local edit at
zero context cost---and the rest of the paper earns that placement. We
defer the broader literature to Section~\ref{sec:related-work}.

\section{Per-user LoRA contaminates globally}
\label{sec:negative}

Weights are an excellent place to keep knowledge---pretraining
packs millions of facts into them accurately, at roughly $2$--$4$
bits per parameter~\citep{allenzhu2025physics}. But that capacity is
\emph{shared} and the facts are \emph{incompressible}: each must
be stored explicitly and ends up entangled with every other---a
distinct share of the weights that does not shrink with
better training~\citep{li2026ikp}. That is fine for the facts
everyone queries. It is a poor
place to keep \emph{one user's} facts written as a \emph{per-user
weight edit}, and not for lack of capacity. A per-user LoRA edits
$Q/K/V/O$ in every layer, so its edit is unconditioned (part of
every forward pass) and must be fit per user from a handful of
examples. Two problems follow, and we measure both: the edit is
global, so it contaminates text that has nothing to do with the
user; and the facts, once trained in, are \emph{recalled but not
readily reasoned over}. The first is an architectural cost paid
on every base; the second matters most for personalization, and
is why a fact belongs in a store the shared reasoning can read
rather than in a per-user copy of the weights.

\subsection{Architectural contamination on the same base (Mini-Engram-d20)}

We hold the base fixed (Mini-Engram-d20@1536, 1.22\,B parameters)
and train a per-user LoRA (rank-64 on Q/K/V, 1{,}500
steps per fact, single-token gold) on the synthetic user fact sets
used throughout. For each of 20 test users we measure val\_bpb on
a held-out 262\,K-token ClimbMix shard---text unrelated to the
user's facts---before and after the edit. Per-user Engram-row
Joint OPT is measured under the identical protocol on the same
base.

\begin{figure}[t]
  \centering
  \includegraphics[width=0.82\linewidth]{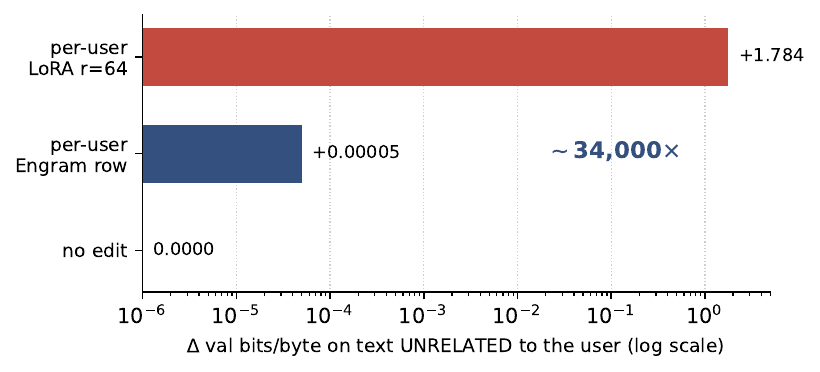}
  \caption{Architectural contamination on held-out text,
  Mini-Engram-d20 (canonical seed S0, $n{=}20$ users). A per-user
  LoRA more than triples val bpb on text \emph{unrelated} to the
  user's facts ($+1.784$; per-user range $[+0.44,+3.74]$; 17/20
  users worse), while a per-user Engram row leaves it unchanged to
  four decimals ($+0.00005$; 0/20 worse)---$\sim$34{,}000$\times$
  less, by design. The 3-seed mean ratio is
  $\sim$33{,}000$\times$ (Appendix~\ref{app:layered-seeds}).}
  \label{fig:contamination}
\end{figure}

Figure~\ref{fig:contamination} shows the gap; the mechanism is the
point. A LoRA
cannot write a fact without bending a function every input
shares---to lift the probability of Maya's cardiologist it moves
$Q/K/V/O$ along directions that also fire on unrelated text---so
the loss on held-out text rises (more than \emph{tripling} here),
and it rises a little further with each fact added, because every
new fact bends the function again. An Engram row is written to an
address instead: it is read only when its trigger N-gram is
hashed, and leaves the loss on everything else unchanged to four
decimal places. This is the difference between editing a function
and writing to a store, and it is precisely what makes
co-locating many users feasible in one and hopeless in the other:
contamination from a shared delta accumulates and cannot be
quarantined per user, whereas addressed writes have zero
cross-talk by construction (Section~\ref{sec:serving}).

\subsection{Recall without reasoning}

A per-user LoRA learns to \emph{recall} its facts---direct recall
is near-perfect---but the harder half of personal memory is
\emph{reasoning over} them, and that is where it struggles. The
reason is composability. Knowledge acquired in pretraining is
laid down in a form the model's reasoning can chain; a per-user
adapter fit to a handful of examples instead memorizes a flat
trigger$\to$answer mapping that the shared reasoning does not
readily pick up and combine. To answer an indirect question the
adapter would have to carry the reasoning skill as well, learned
from the same few examples---hard to train, and it does not
transfer to facts it never saw. This recall-without-reasoning gap
is the recurring finding of the per-user-adapter
line~\citep{su2024prag,tan2024oppu,zhuang2024hydra}: the facts go
in, but indirect questions over them do not reliably come out.

What a user sees in addition depends on the base. We replicate the per-user LoRA recipe (rank-64 NTP on
observation/fact/QA mixtures, 12 epochs $\times$ 200 samples) on
four instruction-tuned bases---Qwen2.5-3B and
Qwen2.5-7B~\citep{qwen2025qwen25},
Llama-3.1-8B~\citep{grattafiori2024llama3}, and
Mistral-7B~\citep{jiang2023mistral}---and one base LM
(Figure~\ref{fig:crossbase}):

\begin{figure}[t]
  \centering
  \includegraphics[width=0.86\linewidth]{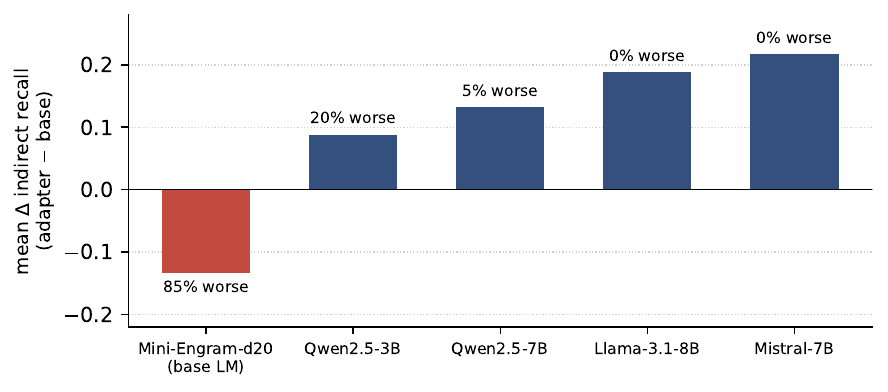}
  \caption{Cross-base LoRA scaling: mean change in indirect recall
  (adapter $-$ base) with the fraction of users it hurts. On the
  base LM (Mini-Engram-d20) a per-user LoRA disrupts fragile
  completion behavior (85\% worse, $\Delta{=}{-}0.133$); on every
  instruction-tuned base the reasoning skill absorbs the
  perturbation (0--20\% worse, $\Delta$ from $+0.088$ to $+0.217$).}
  \label{fig:crossbase}
\end{figure}

The split is clean (Figure~\ref{fig:crossbase}). On a base LM the
global perturbation also disrupts fragile completion behavior,
so indirect recall drops outright; on an instruction-tuned base
the base's own reasoning carries the indirect question and masks
the adapter's inability to provide it. The strength of the base thus
controls whether the damage is visible at all: we can make it appear
or vanish by changing how strong the base is
(Appendix~\ref{sec:layered-sft}). The reasoning has to come from
somewhere, and Section~\ref{sec:layered} supplies it from one shared
skill instead of asking every per-user edit to learn it again.

\section{Method: User as Engram}
\label{sec:method}

\begin{figure}[t]
  \centering
  \includegraphics[width=\linewidth]{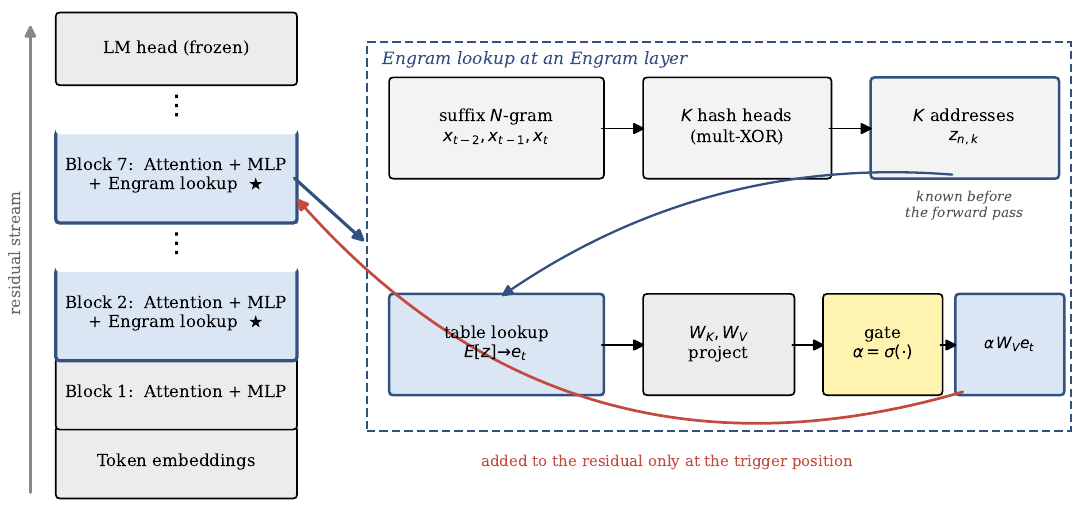}
  \caption{\textbf{Where the Engram lives in the transformer.} The
  backbone is a standard transformer; at a few designated Engram
  layers ($\bigstar$) a content-addressable lookup runs alongside
  attention and the MLP. At a position, the recent tokens' suffix
  $N$-gram is hashed by $K$ heads into $K$ table addresses
  (deterministic, hence known before the forward pass); the
  retrieved rows are projected by $W_K, W_V$, gated by $\alpha$,
  and added to the residual stream---but the gate fires only where
  the trigger $N$-gram is present. Unlike attention and the MLP,
  which touch every position and every weight, the Engram read is
  sparse and addressable; that is what lets a single written row
  change behavior locally.}
  \label{fig:engram-arch}
\end{figure}

\begin{figure}[t]
  \centering
  \includegraphics[width=\linewidth]{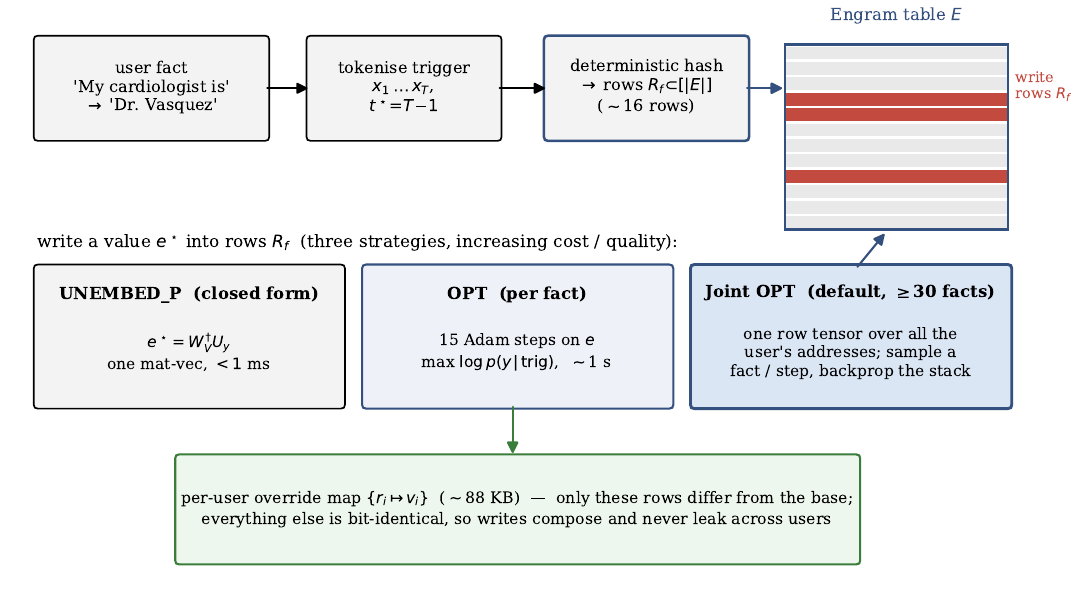}
  \caption{\textbf{Inserting a user fact into the Engram.} A fact
  is reduced to \emph{where} and \emph{what}: the trigger's suffix
  $N$-gram hashes to a sparse set of row addresses $R_f$
  (\emph{where}), and one of three strategies writes a value
  $e^\star$ into those rows (\emph{what}). UNEMBED\_P solves for
  $e^\star$ in closed form; OPT takes a few gradient steps per
  fact; Joint OPT---our default beyond $\sim$30 facts/user---
  optimizes all of the user's rows together so they do not
  interfere. The result is a per-user override map of
  $\sim$88\,KB; every row outside $R_f$, and the entire backbone,
  is left bit-identical.}
  \label{fig:user-insertion}
\end{figure}

\begin{figure}[t]
  \centering
  \includegraphics[width=\linewidth]{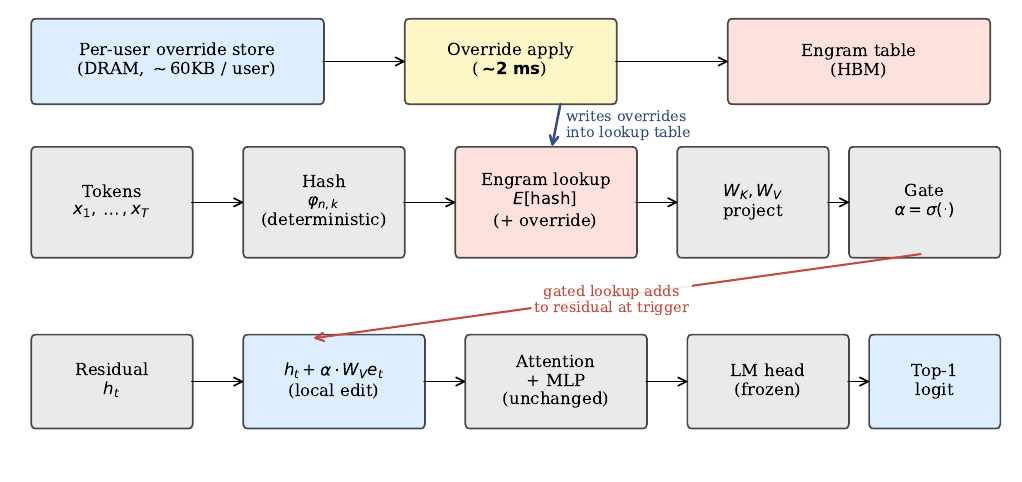}
  \caption{EngramServer architecture. Per user, an override map of
  (row index, row vector) pairs lives in DRAM. On each
  request, the server saves the originals at the affected addresses
  ($\sim$2 ms), writes the user's overrides, runs the forward pass,
  then restores. The Engram lookup at the configured layer
  transparently sees the override values; the gate fires only at the
  trigger N-gram. There is no router and no graph rewrite.}
  \label{fig:arch}
\end{figure}

We instantiate the ``local per-user edit'' concretely: surgically write
per-user fact rows into the hash-keyed memory table of an
Engram-pretrained model. Figure~\ref{fig:engram-arch} shows where that
table sits in the network; the subsections below walk through how a
fact is written into it (Figure~\ref{fig:user-insertion}) and
how the writes are served per user (Figure~\ref{fig:arch}).

\subsection{Where the Engram lives}
\label{sec:engram-arch-text}

Most of an Engram model is an ordinary
transformer~\citep{vaswani2017transformer}; the only addition is
a content-addressable lookup spliced into a few layers
(Figure~\ref{fig:engram-arch}; two layers in our Mini-Engrams).
At every position, the recent tokens---a suffix $N$-gram---are
hashed by $K$ deterministic multiplicative-XOR heads into
addresses in a large embedding table $E$. The retrieved rows are
projected by $W_K, W_V$, weighted by an attention-style gate
$\alpha=\sigma(\cdot)$, and \emph{added to the residual stream},
exactly where attention and the MLP also write. The lookup is
thus a read from memory, but keyed by the surface form of the
recent tokens rather than by query--key similarity across the
sequence.

Two properties make this a genuine store rather than just more
parameters, and both are what we exploit. First, the address is a
deterministic function of the input token IDs---known
\emph{before} the forward pass---so we know exactly which rows a
given query will read, and the table can be offloaded to host
DRAM. Second, the gate makes the read sparse and conditional:
most positions retrieve essentially nothing, and a stored row
affects the output only at positions whose $N$-gram addresses it.
This is the addressable, sparse store that per-user facts
want---the opposite of the global weight edit of
Section~\ref{sec:negative}.
Attention and the MLP are dense and global---every position,
every weight---whereas the Engram read touches a handful of rows
at a handful of positions, so editing one row changes the model
only where that row is addressed. The rest of this section turns
that into a per-user memory.

\subsection{Surgical row insertion}

To write a fact, we put its answer where its trigger is read. The
trigger's suffix $N$-gram hashes to a fixed, sparse set of table
rows $R_f$---about 16 rows for our $N{=}3$, $K{=}8$
configuration---deterministically, touching no other row. What
remains is \emph{what} to write into those rows, for which we use
three strategies of increasing cost and fidelity (per-strategy
recall against a RANDOM control in Appendix~\ref{app:strategies}):
\begin{itemize}[leftmargin=*,topsep=1pt,itemsep=1pt]
  \item \textbf{UNEMBED\_P} (closed form): solve in one step for
        the row value whose projection steers the next-token
        distribution toward the answer---a single matrix--vector
        product, ${<}1$\,ms, no training.
  \item \textbf{OPT}: refine that value with a few gradient steps
        per fact ($\sim$1\,s), trading time for accuracy.
  \item \textbf{Joint OPT} (our default beyond $\sim$30
        facts/user): optimize \emph{all} of a user's rows
        together rather than one fact at a time (full algorithm
        in Appendix~\ref{app:joint-opt}; convergence behavior in
        Appendix~\ref{app:joint-loss}).
\end{itemize}
Joint OPT is the default because of the one place this design has
to earn its keep. When many of a user's facts are live at once,
their rows are read in the same forward passes and interfere;
optimizing them jointly lets the rows settle into values that
coexist, which is what holds recall up as facts-per-user grows
(Section~\ref{sec:experiments}). Crucially, every write still
lands only in that user's $R_f$, so however many facts or domains
are stacked, the store stays additive and private---the locality
is the architecture's, not a constraint we have to impose.

Read together, the two figures give the method in one sentence:
\textbf{remembering a fact is writing a few rows.} A fact
decomposes into an address set $R_f$ (\emph{where}, fixed by the
trigger's hash) and a value $e^\star$ (\emph{what}, set by one of
the strategies above), and the write touches nothing else. The
locality we rely on is therefore \emph{inherited from the
architecture}---the read was already sparse and addressed
(Section~\ref{sec:engram-arch-text})---rather than enforced by a
separate mechanism. That is why per-user and per-domain writes
compose without a combiner and never leak across users
(Section~\ref{sec:experiments}), and Section~\ref{sec:locality}
verifies that ``touches nothing else'' is literally exact.

\subsection{The local edit is a glass box}
\label{sec:locality}

\begin{figure}[t]
  \centering
  \includegraphics[width=\linewidth]{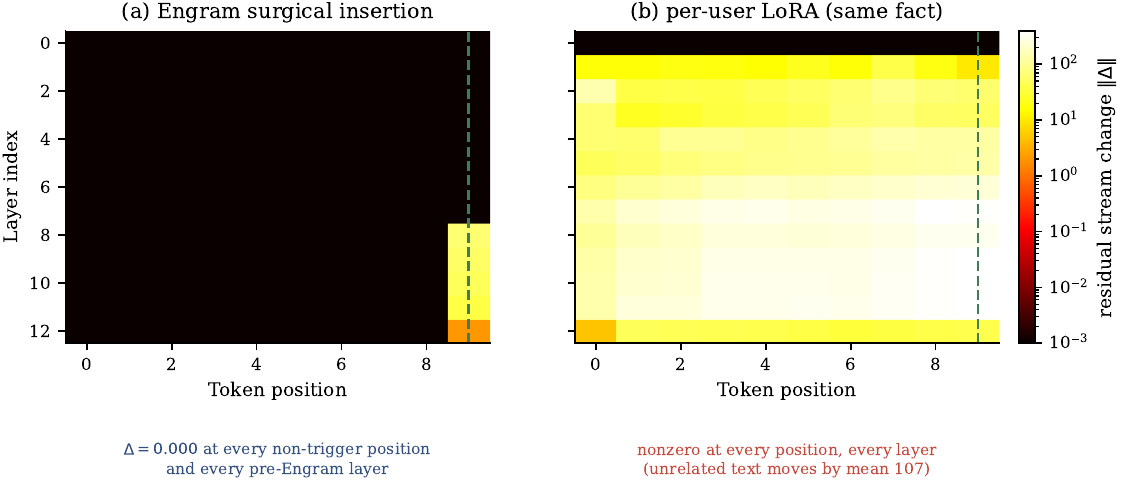}
  \caption{\textbf{Addressed write vs.\ global function-bend.}
  Per-layer, per-position residual-stream change
  $\|x^{(\ell)}_{\text{after}}-x^{(\ell)}_{\text{before}}\|$ on the
  \emph{same} trigger sentence and base (Mini-Engram-d12@1280, log
  color scale). \textbf{(a)} An Engram row insertion is
  \emph{exactly} 0.000 at every position before the Engram layer
  (causality) and at every non-trigger position after it---only the
  trigger column moves. \textbf{(b)} A per-user LoRA fit to the
  \emph{same single fact} moves the residual at every position and
  every layer, and perturbs unrelated text (``The capital of France
  is'') by mean $107$. Storing a fact in a shared function changes
  the function everywhere; storing it at an address does not.}
  \label{fig:locality}
\end{figure}

\begin{figure}[t]
  \centering
  \includegraphics[width=\linewidth]{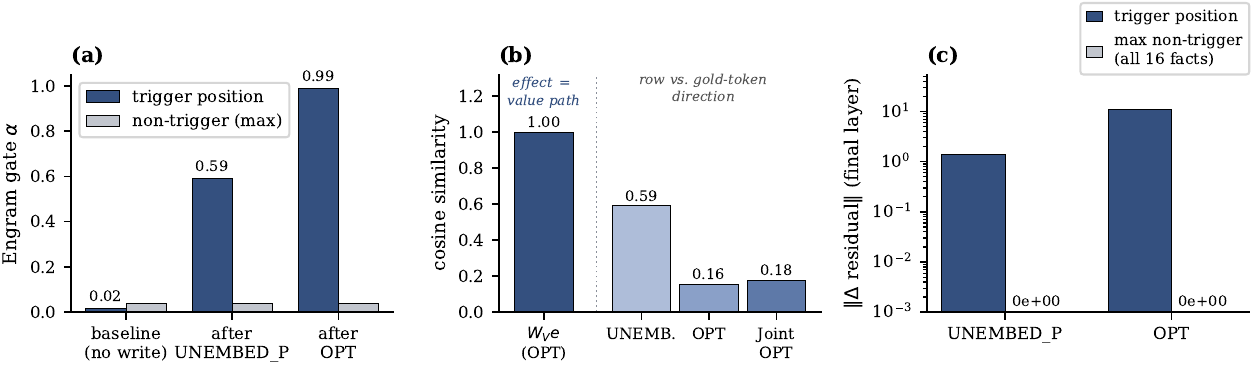}
  \caption{The mechanism on the trained model (Mini-Engram-d20,
  16 facts). \textbf{(a)} The write opens its own gate: the
  trigger-position gate $\alpha$ rises from $0.02$ to $0.99$ (OPT)
  while non-trigger positions stay at $0.04$. \textbf{(b)} The
  deployed row's residual change is cosine $0.999$ to its value-path
  projection $W_V e$ (left bar); what the row \emph{encodes} relative
  to the gold token's unembedding is exact for the closed-form
  solution ($0.59$) and drifts as gradient OPT/Joint-OPT trade direct
  gold-alignment for higher recall ($0.16$--$0.18$). \textbf{(c)}
  Exact locality: the maximum non-trigger residual change over all 16
  facts is $0.000$ for both strategies, versus an order-1 change at
  the trigger.}
  \label{fig:glassbox}
\end{figure}

A LoRA's change is spread invisibly across the whole model, so you
cannot point to what it did. An Engram write is the opposite: you can
watch every step of it on the trained model. We follow one inserted
fact through the write and check each step on the trained
Mini-Engrams (full detail, with the depth and multi-hop probes, in
Appendix~\ref{app:mech}). Three things are worth seeing, and
Figures~\ref{fig:locality} and~\ref{fig:glassbox} show each.

\paragraph{(1) The write switches on its own lookup, and adds just
the value it should.} A row reaches the output only through a gated
value, $\alpha_t\,W_V e$: a switch $\alpha_t$ that decides whether the
lookup is read, times the value $W_V e$ the row carries. Writing a
fact does both at once. The switch at the trigger turns from nearly
off to nearly on, while it stays off everywhere else
(Figure~\ref{fig:glassbox}a)---the row we write is also what turns its
own lookup on. And the change the row makes to the output points
almost exactly along the value it carries (Figure~\ref{fig:glassbox}b):
the switch and the small convolution around it scale the value but do
not turn it into something else. The same alignment was reported for
the original Engram code on a fresh, untrained model; we confirm it on
the trained model, for the strategy we actually deploy.

\paragraph{(2) Nothing else moves---exactly nothing.} We write only
the trigger's rows, and the trigger's last few tokens are unique in
the sentence, so every other position reads the rows it always read
and its output is unchanged to the last bit. This holds for every one
of the 16 test facts and both writing strategies: the largest change
at any non-trigger position, at any layer, is $0$
(Figure~\ref{fig:locality}a). A LoRA that learns the same fact does
the opposite---it moves every position in every layer, and shifts
unrelated text as well (Figure~\ref{fig:locality}b). This is the line
between editing a function and writing to a store. Note where the
locality comes from: not from the switch being off elsewhere, but from
the fact that we only ever changed the rows at one address.

\paragraph{(3) It only works near the end of the network.} The Engram
lookup is read at a late layer, after the model has mostly made up its
mind about the next token. That is why a row can change the answer
cleanly. To check this is the real cause and not a coincidence, we
write the same facts into the model's \emph{early} lookup instead: at
the same effort, recall drops from perfect to about a quarter, because
an edit made early gets reworked by all the layers above it. Where the
lookup sits is not a free choice---it has to be late enough that the
value it adds is close to the final answer.

What this kind of write \emph{cannot} do is join two facts together.
The lookup only matches the words it was given, so it answers ``who is
my doctor'' but not ``where does my doctor work'' unless that second
step was stored too (Section~\ref{sec:multihop}). That is the reason we
keep the facts and the reasoning apart (Section~\ref{sec:layered}): the
store can hold a fact but cannot chain it, so the chaining has to come
from somewhere else. The split between facts and reasoning is forced by
how the mechanism works, not a convenience we chose.

\subsection{Per-user override tables and additive composition}

The production design uses per-user override tables. Each user
has a small dictionary $\{r_i \mapsto v_i\}$ of fact rows. At
request time, the server saves the originals at the affected
addresses, writes the user's overrides, runs the forward, and
restores. Cross-user leakage is \emph{zero by design}---%
user A's overrides are not in the table when user B queries. (We
also benchmarked a shared table with a per-user hash salt and
rejected it for lower recall; Appendix~\ref{app:alt}.)

Override maps with disjoint addresses commute, so corporate
facts + user facts (or any number of domain-specific Engrams)
\textbf{stack additively at inference} without retraining a
combiner. This mirrors how Stable Diffusion LoRAs additively
stack, but at the row level. Section~\ref{sec:experiments}
confirms that additive composition is lossless when the domains
have disjoint trigger templates.

\section{Experiments}
\label{sec:experiments}

Before composing the store with a shared reasoning skill
(Section~\ref{sec:layered}), we first characterize the bare
per-user store on its own---how well facts surface, what they cost,
and where it breaks---so that the layered design's gains can be
attributed cleanly to the skill it adds.
This section answers four questions about User-as-Engram, in order.
\textbf{(a)~How accurately do facts surface?}
(Section~\ref{sec:density}) for one fact and for many at once.
\textbf{(b)~What does it cost?} (Section~\ref{sec:storage}) storage
and composition vs.\ a per-user LoRA. \textbf{(c)~Does writing the
fact beat retrieving it?} (Sections~\ref{sec:memory-systems}--\ref{sec:locomo})
against memory-system and RAG baselines on the same answering LM,
including paraphrased queries. \textbf{(d)~Where does it break?}
(Section~\ref{sec:multihop}) multi-hop chaining---the store
matches words but does not compose them. The capacity, token-budget, and dense-size
scaling ablations that justify our recipe are in
Appendix~\ref{app:extended}.
We use three fact corpora throughout: the \textbf{base} 200-fact
benchmark (100 USER + 100 ORG facts), the \textbf{XL} corpus of
1{,}000 USER + 1{,}000 ORG templated facts
(Section~\ref{sec:factscale}), and the \textbf{XXL} corpus of
3{,}132 distinct trigger templates (high-density and
distinct-template stress tests).

\subsection{Setup: the Mini-Engram models we insert into}
\label{sec:repro}

Before measuring per-user insertion we need an Engram backbone to
insert into. \citet{cheng2026engram} released only architectural reference
code; no Engram-trained weights are public. We graft Engram into
Karpathy's nanochat~\citep{karpathy2026nanochat}---a GPT-2-style
backbone~\citep{radford2019gpt2} optimized with
Muon+AdamW~\citep{jordan2024muon}---and train four Mini-Engram
models on a single Blackwell GPU at the ablation-optimal recipe
(Section~\ref{sec:capacity-ablation}; pretraining curves in
Appendix~\ref{app:pretrain}): the same \texttt{large}
Engram table (50\,K $\times$ 256, 51.2\,M params), trained at
Karpathy's 12 tokens-per-param (t/p) budget on the scaling-params
for every dense size.

\paragraph{Model-name notation.} Throughout the paper we use
\texttt{d\{depth\}@\{width\}} for the ablation-optimal Mini-Engrams
(\texttt{d12@768}, \texttt{d12@1280}, \texttt{d20@1536}) and
\texttt{v1}/\texttt{v2} to denote earlier and final training runs
at the same shape. Where the width is omitted (e.g.\ \texttt{d12}
without a suffix), we mean the \texttt{v2} (ablation-optimal at
width 768, 339\,M total) checkpoint. The four headline checkpoints,
their param counts and token budgets are in
Table~\ref{tab:miniengram}; all four share the same 51.2\,M Engram
table.

\begin{table}[t]
  \centering
  \caption{Mini-Engram pretraining at the ablation-optimal recipe.
  All use the \texttt{large} Engram (50\,K $\times$ 256 = 51.2\,M
  Engram params). Trained on a single NVIDIA RTX PRO 6000 Blackwell
  (102\,GB) in bf16.}
  \label{tab:miniengram}
  \begin{tabular}{lcccc}
    \toprule
     & d8 v2 & d12 v2 & \textbf{d12@1280 opt.} & \textbf{d20@1536 opt.} \\
    \midrule
    depth $\times$ width   & 8 $\times$ 512  & 12 $\times$ 768 & 12 $\times$ 1280 & 20 $\times$ 1536 \\
    Engram layers          & 2, 5            & 2, 7            & 2, 7             & 2, 11 \\
    total params           & 178\,M          & 339\,M          & 625\,M           & \textbf{1.22\,B} \\
    scaling-params         & 42\,M           & 110\,M          & 278\,M           & 617\,M \\
    ClimbMix tokens (12 t/p)& 0.50\,B         & 1.32\,B         & 3.34\,B          & 7.40\,B \\
    final val\_bpb         & 0.912           & 0.827           & 0.770            & \textbf{0.730} \\
    wall-clock (single GPU)& $\sim$50\,min   & $\sim$5\,h      & $\sim$10.5\,h    & $\sim$70\,h (shared GPU) \\
    \bottomrule
  \end{tabular}
\end{table}

\paragraph{We reproduce Engram's two signature findings.}
Suppressing the Engram lookup hurts factual recall far more than
reading comprehension---the factual-vs-reading sensitivity
asymmetry of \citet{cheng2026engram} (their \S6.3), at the
$\sim$10$\times$ smaller magnitude expected for a model two
orders smaller on a corpus three orders smaller. The Engram
layers also act as effective extra depth: a LogitLens probe shows
our small Engram model resolving its prediction earlier than its
dense-only twin, matching their \S6.1 (Appendix
Figure~\ref{fig:logitlens}).

\subsection{Per-fact recall and within-user density}
\label{sec:density}
\label{sec:single-fact}

\begin{figure}[t]
  \centering
  \includegraphics[width=\linewidth]{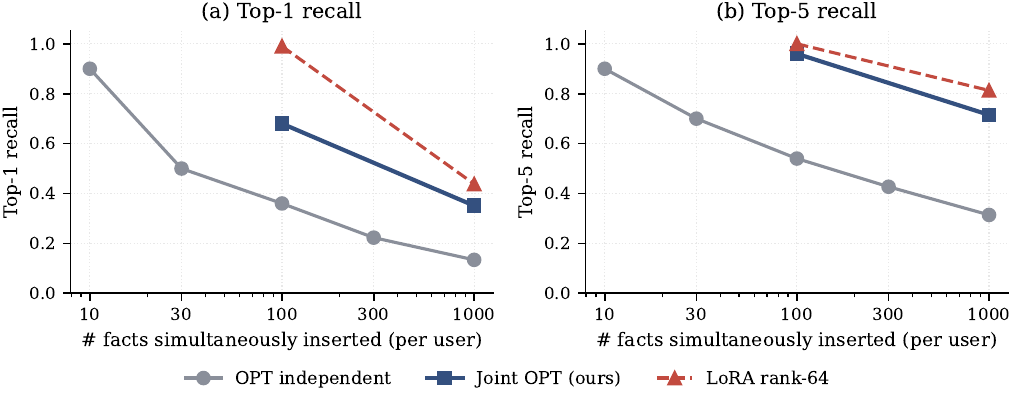}
  \caption{Within-user fact-density: top-1 (left) and top-5 (right)
  recall vs.\ number of facts inserted simultaneously into one user's
  override map. Joint OPT (blue) closes most of the gap to LoRA
  rank-64 (red dashed) at 161$\times$ less storage (88\,KB vs.\
  14.2\,MB at 100 facts).}
  \label{fig:density}
\end{figure}

A single inserted fact surfaces within $\sim$8 points of the
in-context ceiling, at three orders of magnitude lower per-user
cost than training a full LoRA (Appendix~\ref{app:strategies}). The
harder question is what happens when many facts are live at
once: the rows in a single user's override table interfere during
the forward pass. We sweep that density curve on Mini-Engram-d12
(Figure~\ref{fig:density}).

\textbf{Joint OPT is the recommended default for $\geq$30
facts/user}: it tracks multi-fact LoRA rank-64 to within a few
top-5 points while keeping top-5 recall above 90\% out to
$\sim$300 facts/user, all at up to $\sim$161$\times$ less
per-user storage (Figure~\ref{fig:density}; per-$n$ recall and
storage in Appendix~\ref{app:extended}).

\paragraph{What the density ceiling is---and what it is not.} Three
ablations triangulate the ceiling to a single cause, ruling out the
obvious suspects. (i) It is \emph{not} backbone capacity: holding
Engram capacity fixed and scaling the dense backbone does not lift
recall at 1\,000 facts---it slightly degrades it (top-1
$0.35\!\to\!0.28$ from d12@768 to d20@1536,
Section~\ref{sec:future}). (ii) It is \emph{not} the table or the
hash: per-fact \emph{independent} OPT, where each fact gets private
rows, shows \emph{no} ceiling to 1\,000 facts ($\geq0.98$ top-1,
Appendix~\ref{sec:factscale}). (iii) It is \emph{not} pretraining: a
multi-fact-in-the-loss finetune only accelerates convergence---given
enough OPT steps the baseline reaches the same asymptote
(Section~\ref{sec:mf-finetune}). What remains is the one thing all
three share: \textbf{gradient interference among the co-active rows
in a single user's table, at a fixed inference-time optimization
budget}. This is why the path to breaking it is a recipe change
(multi-fact-in-the-loss, scoped tables), not further dense scaling,
and why it scales with facts-per-user rather than with the model.

\subsection{Cost, storage, and composition}
\label{sec:storage}
\label{sec:method-cmp}
\label{sec:additive}

\begin{figure}[t]
  \centering
  \includegraphics[width=0.78\linewidth]{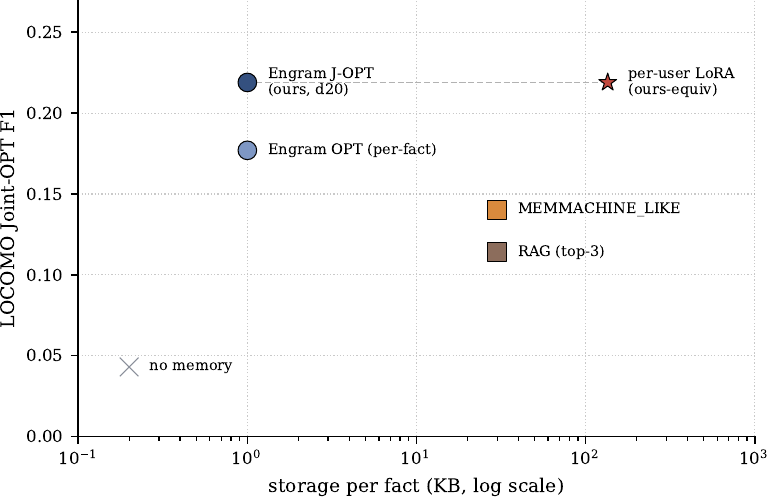}
  \caption{Cost-quality trade-off across personal-memory methods
  at the same answering LM (Mini-Engram-d20@1536). User-as-Engram
  Joint OPT matches a per-user LoRA's LOCOMO F1 at
  161$\times$ smaller per-user storage (88\,KB vs.\ 14.2\,MB at
  100 facts). Retrieval baselines store each fact at a higher
  per-fact cost than an Engram row and still cap out below Engram
  on quality.}
  \label{fig:pareto}
\end{figure}

\textbf{Storage.} A user's whole table is $\sim$88\,KB at 100
facts/user and grows about linearly with the fact count
($\sim$0.9\,KB/fact), while a per-user LoRA is a
fact-count-independent 14.2\,MB---$\sim$161$\times$ larger at 100
facts/user, and $\sim$1700$\times$ at 10 facts/user. At a million
users that is 100\,GB versus 14.2\,TB: the difference between one
server and a distributed store (Appendix Figure~\ref{fig:scaling}).
Figure~\ref{fig:pareto} places all the methods on one cost--quality
plot: a single fact matches the in-context ceiling at a fraction of
LoRA's cost, and at high density Joint OPT keeps pace with even
rank-64 LoRA at tens of times less storage, with the gap turning in
Engram's favor by 1\,000 facts.

\textbf{Composition.} Two users' tables, or a user's facts and a
company's, occupy different addresses, so they simply add up---no
combiner to train. The one exception is when two sets of facts share
the same trigger words: their addresses collide and the last write
wins. The rule, then, is to add tables when their triggers differ and
to keep separate per-user tables when they do not
(Appendix~\ref{app:multidomain}).

\subsection{Comparison against memory systems}
\label{sec:memory-systems}

\begin{figure}[t]
  \centering
  \begin{subfigure}{0.49\linewidth}
    \centering
    \includegraphics[width=\linewidth]{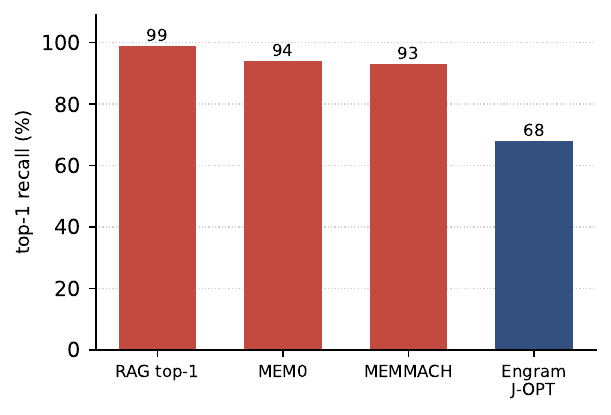}
    \caption{Retrieval wins on exact-trigger queries---but Engram
    pays \emph{zero} context tokens (0 vs.\ 16--63) for its 68\%.}
    \label{fig:memsystems-a}
  \end{subfigure}\hfill
  \begin{subfigure}{0.49\linewidth}
    \centering
    \includegraphics[width=\linewidth]{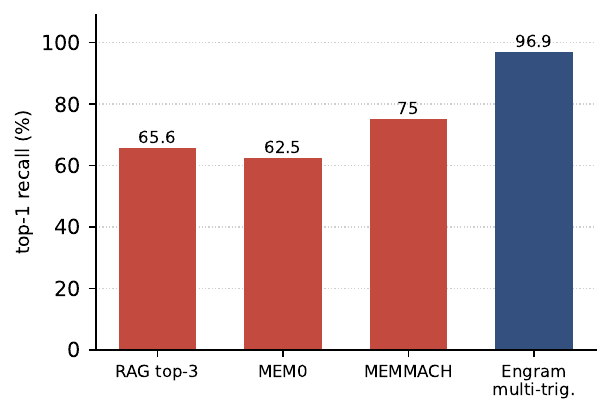}
    \caption{Engram multi-trigger insertion wins on paraphrases by
    22 points, again at zero context.}
    \label{fig:memsystems-b}
  \end{subfigure}
  \caption{Direct recall vs.\ memory systems, all sharing the same
  Mini-Engram-d12 answering LM (100 USER facts, XXL corpus).
  \subref{fig:memsystems-a}~When the query is the fact's stored
  prefix, nearest-neighbor retrieval is near-perfect and beats
  Engram's 68\% top-1---but at a context-token cost Engram does not
  pay. \subref{fig:memsystems-b}~When the query is a paraphrase,
  retrieval drops to 60--75\% while multi-trigger Engram insertion
  reaches 96.9\% top-1.}
  \label{fig:memsystems}
\end{figure}

We compare User-as-Engram against state-of-the-art memory
systems that target the same use case (per-user / personal
memory). All systems use the \emph{same} Mini-Engram-d12 as the
final-answer LM; the only difference is how facts are stored
and retrieved. Our \texttt{MEM0} and \texttt{MEMMACHINE} baselines
follow the retrieval recipes of Mem0~\citep{chhikara2025mem0} and
MemMachine~\citep{wang2026memmachine}. The sentence encoder used
by the RAG / MEM0 / MEMMACHINE baselines is
\texttt{all-MiniLM-L6-v2}~\citep{wang2020minilm,reimers2019sbert}
(80\,M parameters; comparable in size to a production retriever).

This subsection probes \emph{direct} fact recall under
trigger-exact and paraphrased queries; the separate
\emph{indirect-reasoning} comparison (does retrieval recover
the gold fact across hops, and what happens as the KB grows?)
is deferred to
Sections~\ref{sec:layered-rag}--\ref{sec:rag-scale}.

\paragraph{Setup.} 100 USER facts (XXL corpus), asked two ways: with
the fact's exact stored prompt (easy for retrieval---the words match),
and with a paraphrase (``I love the spice'' for ``My favorite spice
is''; 16 facts $\times$ 4 rewordings).

On exact-trigger queries the words line up with what was stored, so
nearest-neighbor retrieval is near-perfect and beats Engram's 68\%
top-1---but it pays 16--63 context tokens for the answer, where Engram
pays none (Figure~\ref{fig:memsystems}).

\textbf{The story flips on paraphrased queries.} When the query
is reworded, the sentence encoder confuses surface forms and
retrieval slips, while a single-trigger Engram misses the
reworded N-gram entirely. The fix is to write a row at every
anticipated paraphrase: multi-trigger insertion then wins on
paraphrase by a clear margin at zero context cost
(Figure~\ref{fig:memsystems}), because the contract is
explicit---the trigger N-gram---rather than left to an encoder's
notion of similarity.

\paragraph{Scale invariance.} This verdict (\emph{retrieval wins on
the verbatim prefix; multi-trigger Engram wins under paraphrase}) is
invariant to dense scale: the Engram numbers barely move across
our three sizes, and only the retrieval baselines drift as their
answering-LM improves (Appendix~\ref{app:extended}).

Our retrieval baselines are the retrieval step itself (top-3
sentence-encoder lookup over atomic facts), not the full Mem0 or
MemMachine systems with their extraction and consolidation
machinery; we skip those because our facts are already
structured~\citep{maharana2024locomo,wu2024longmemeval}.

\paragraph{Paraphrase generalization.}
\label{sec:paraphrase}
The mechanism behind the paraphrase result is worth stating, because
it is also a limit: the hash addresses a token-level suffix
$N$-gram, so different surface forms of the same fact map to
\emph{different} rows. A fact written once therefore transfers to a
paraphrase only when the paraphrase ends in the same tokens; writing
the fact at every anticipated phrasing (multi-trigger insertion) is
what closes the gap and drives the $96.9\%$ vs.\ $75.0\%$ win above.
The contract is explicit---the trigger $N$-gram---rather than left to
an encoder's notion of similarity. Single- vs.\ multi-trigger
generalization is detailed in Appendix~\ref{app:paraphrase}, with
per-paraphrase rank data in Appendix~\ref{app:paraphrase-detail}.

\subsection{LOCOMO single-hop fact recall}
\label{sec:locomo}

To evaluate the same memory systems on a published long-term
conversational benchmark, we run all eight systems from
Section~\ref{sec:memory-systems} on
LOCOMO~\citep{maharana2024locomo}. We use \emph{all 10 LOCOMO
conversations} and take the first 80 single-hop (non-adversarial)
questions per conversation, for a total of $\sim$800
question-answer pairs grounded in dialog per model.

\paragraph{Setup.} Retrieval baselines store the gold evidence turns;
User-as-Engram stores the gold (question, answer) pair. Every system
uses the same Mini-Engram-d12 to write the final answer, so the only
difference is how the answer is found. We score token-F1 against the
gold answer (details in Appendix~\ref{app:locomo}).

\begin{figure}[t]
  \centering
  \includegraphics[width=\linewidth]{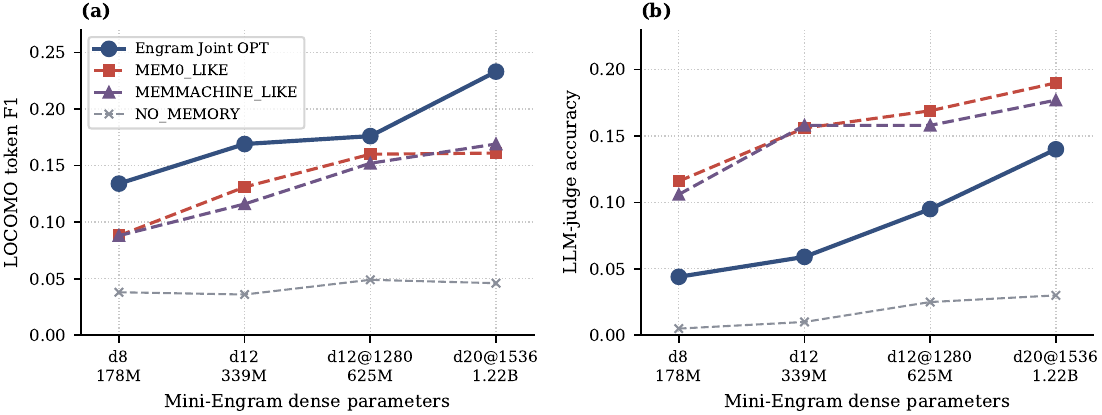}
  \caption{LOCOMO single-hop, full 10 conversations, scaling with
  Mini-Engram dense size. \textbf{(a)} \emph{Token-F1}:
  User-as-Engram Joint OPT (blue) beats every retrieval baseline
  at every scale. \textbf{(b)} \emph{LLM-judge accuracy}
  (Qwen2.5-14B-Instruct): the ranking flips---retrieval baselines
  (MEM0\_LIKE, MEMMACHINE) beat Joint OPT by 0.05--0.10 because
  token-F1 over-credits Engram for the correct first answer
  token despite a noisy continuation.}
  \label{fig:locomo-scaling}
\end{figure}

\textbf{User-as-Engram Joint OPT clears every retrieval baseline
on single-hop token-F1} (Figure~\ref{fig:locomo-scaling}).
Absolute scores are low---Mini-Engram-d12 is a small base LM with
no instruction tuning and token-F1 punishes verbosity---so the
relative ordering is what matters. This is a best-case-for-each
benchmark: retrieval stores the gold evidence sentence, Engram
stores the gold $(q, a)$ pair, and the only question is which
extracts the answer more reliably under a fixed LM. Engram wins
because the gate fires on the question's trigger N-gram and
biases the next token toward the gold answer, whereas retrieval
can miss the relevant turn and, even when it hits, the
in-context evidence still competes with the base LM's priors.

\paragraph{Scaling at the optimal config.} We repeated the same
LOCOMO setup with all four trained Mini-Engrams at our
ablation-optimal recipe (Section~\ref{sec:capacity-ablation});
the Engram lead widens monotonically with dense size
(Figure~\ref{fig:locomo-scaling}).

\begin{figure}[t]
  \centering
  \includegraphics[width=\linewidth]{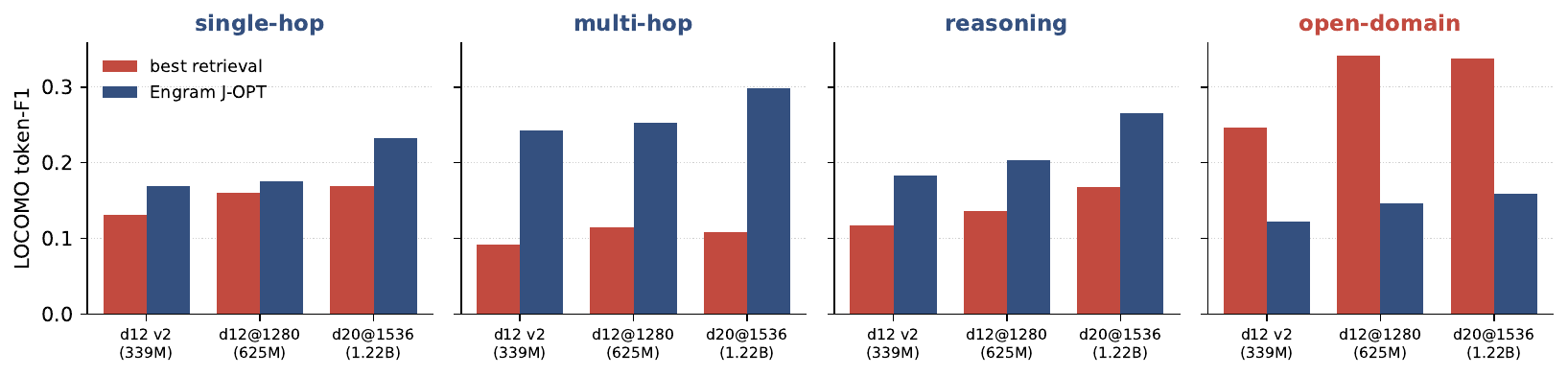}
  \caption{LOCOMO category breakdown (token-F1), Engram Joint OPT
  (blue) vs.\ the best retrieval baseline (red; max of
  MEM0\_LIKE and MEMMACHINE\_LIKE) across three dense scales.
  Engram wins single-hop, multi-hop, and reasoning at every
  scale---the per-fact (question, answer) loss encodes the
  question$\to$answer map that retrieval must chain across
  evidence sentences---but loses open-domain, where the answer is
  a verbatim span of an unseen evidence sentence that a
  single-token bias cannot reconstruct.}
  \label{fig:locomo-categories}
\end{figure}

\textbf{But the win depends on the kind of question.} Across
LOCOMO's four answerable categories, writing the fact wins on
single-hop, multi-hop, and reasoning questions---where the
stored question$\to$answer pair is exactly what is needed---but
loses on open-domain questions, where the answer is a long verbatim
span of a sentence the model never saw and a one-token nudge cannot
rebuild it (Figure~\ref{fig:locomo-categories} and
Table~\ref{tab:locomo-categories}).
So User-as-Engram is a specialized tool, not a drop-in replacement
for retrieval: a real system would route by question type, or use
Engram for the facts it was taught and retrieval for free-form
answers from the session.

\begin{table}[t]
  \centering
  \caption{LOCOMO category $\times$ model token-F1. Engram Joint OPT vs.\
  the best retrieval baseline (MEM0\_LIKE or MEMMACHINE\_LIKE)
  per cell. The Engram win is category-dependent
  (Figure~\ref{fig:locomo-categories}).}
  \label{tab:locomo-categories}
  \resizebox{\linewidth}{!}{
  \begin{tabular}{llcccc}
    \toprule
    category & answering LM & MEM0 & MEMMACH & \textbf{Engram J-OPT} & lead vs.\ best retr. \\
    \midrule
    \multirow{3}{*}{single-hop}  & d12 v2   & 0.131 & 0.116 & \textbf{0.169} & +29\% \\
                                  & d12@1280 & 0.160 & 0.152 & \textbf{0.176} & +10\% \\
                                  & d20      & 0.161 & 0.169 & \textbf{0.233} & +38\% \\
    \midrule
    \multirow{3}{*}{multi-hop}   & d12 v2   & 0.092 & 0.076 & \textbf{0.243} & \textbf{+165\%} \\
                                  & d12@1280 & 0.115 & 0.104 & \textbf{0.253} & \textbf{+120\%} \\
                                  & d20      & 0.102 & 0.108 & \textbf{0.299} & \textbf{+178\%} \\
    \midrule
    \multirow{3}{*}{reasoning}   & d12 v2   & 0.117 & 0.113 & \textbf{0.183} & +56\% \\
                                  & d12@1280 & 0.123 & 0.136 & \textbf{0.203} & +49\% \\
                                  & d20      & 0.135 & 0.168 & \textbf{0.266} & +58\% \\
    \midrule
    \multirow{3}{*}{\textbf{open-domain}} & d12 v2   & \textbf{0.247} & 0.209 & 0.122 & \textbf{$-$51\%} \\
                                  & d12@1280 & \textbf{0.341} & 0.327 & 0.146 & \textbf{$-$57\%} \\
                                  & d20      & 0.314 & \textbf{0.338} & 0.159 & \textbf{$-$53\%} \\
    \bottomrule
  \end{tabular}}
\end{table}

\paragraph{One caveat about the score, and its fix.} Token-F1 gives
credit for partial word overlap, and our basic insertion only trains
the \emph{first} answer token, so it can score well by getting that
first word right over an otherwise noisy answer. When we re-score
with an LLM judge~\citep{zheng2023mtbench} (Qwen2.5-14B,
Appendix~\ref{app:locomo}), first-token
Engram indeed slips behind retrieval on single-hop---but training the
\emph{whole} answer with multi-token Joint OPT fixes it and again
overtakes the best retrieval baseline on the larger models. The
paraphrase and storage wins above score only the first token, so they
are unaffected. (The judge is stricter than LOCOMO's official one, so
these numbers are not directly comparable to its leaderboard.)

\subsection{Multi-hop reasoning over inserted facts}
\label{sec:multihop}

We probe chaining directly: store two facts (``my doctor is
Patel'', ``Patel works at Globex'') and ask a question that needs
both (``where does my doctor work?''). On 63 such pairs, half worded
so the question ends in the same words as the second stored fact and
half not, we can separate a true chain from a lucky word-match.

\begin{figure}[t]
  \centering
  \includegraphics[width=0.72\linewidth]{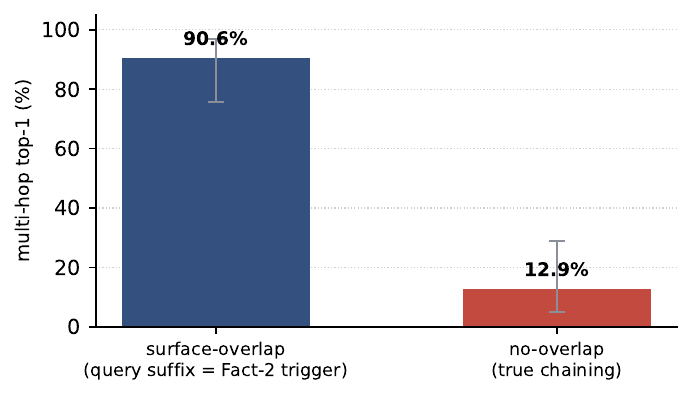}
  \caption{Multi-hop reasoning over Engram-inserted facts
  ($n{=}63$ chained pairs on Mini-Engram-d20, two OPT-15
  insertions per item). Per-fact direct recall is 99.2\% (both
  facts are insertable), so the gap is chaining, not insertion.
  When the query's suffix N-gram coincides with Fact-2's trigger,
  the gate fires and recall is 90.6\%; when true chaining is
  required, it collapses to 12.9\% (Wilson 95\% CIs shown). The
  gate is a surface-N-gram hash---it matches, it does not compose.}
  \label{fig:multihop}
\end{figure}

The result is almost all-or-nothing (Figure~\ref{fig:multihop}): when
the question happens to end in the second fact's words, recall is
high; when a real chain is needed, it falls to near chance---and the
few hits are word-match coincidences. Since each fact alone is
recalled essentially perfectly, the gap is chaining, not storage.
This is the limit the store shares with a per-user LoRA: it matches
words, it does not reason across them. The recite-then-reason recipe
(Section~\ref{sec:discussion}) is the natural fix.

\section{Layered architecture: shared reasoning skill + per-user content}
\label{sec:layered}

So far: per-user LoRA contaminates (Section~\ref{sec:negative})
and per-user Engram preserves direct recall but lacks in-context
reasoning skill (the per-user Engram alone reaches only 23\%
indirect\_any at $n{=}20$ in
Figure~\ref{fig:layered-conditions}). Can the two be \emph{composed}
without inheriting LoRA's damage? Our thesis: personal memory
splits into \emph{content} (per-user, trigger-keyable,
low-cost) and \emph{reasoning skill} (shared patterns,
learnable from held-out users); storing each the right way beats
every all-in-one baseline on every measure. Two
hypotheses make this concrete:
\begin{itemize}[leftmargin=*,topsep=2pt,itemsep=1pt]
  \item \textbf{H1 (naive combination)}: the naive stack of
        per-user LoRA $+$ per-user Engram inherits
        LoRA's indirect-reasoning damage; Engram's local edit does
        not rescue the composition.
  \item \textbf{H2 (layered architecture)}: one \emph{shared}
        LoRA (cross-user reasoning skill) $+$ per-user Engram (local
        content), the layered design, splits memory correctly. The
        shared LoRA's contamination is paid once and shared across
        all users, and the per-user Engram adds none on top (the
        layered design's $\Delta$bpb equals the shared LoRA's alone,
        by Section~\ref{sec:method}'s locality property).
\end{itemize}

\subsection{Design}

Train a single shared LoRA on cross-user in-context-reasoning data:
for each held-out training user (here, u020–u029), render each
indirect QA as a completion-format sample
``\texttt{Facts: <fact1>. <fact2>. ... Q: <q> A: <gold>}'', using only
the facts the question needs. This teaches the LoRA the
\emph{pattern} ``given facts in context, do the reasoning,'' without
memorizing any specific user's content. At inference, attach the
shared LoRA + the test user's Engram-row override and ask the indirect
question without facts in the prompt: the Engram provides content (via
trigger N-gram lookup), the shared LoRA provides the skill
(Figure~\ref{fig:layered-arch}).

\subsection{Six-condition head-to-head}

\begin{figure}[t]
  \centering
  \includegraphics[width=\linewidth]{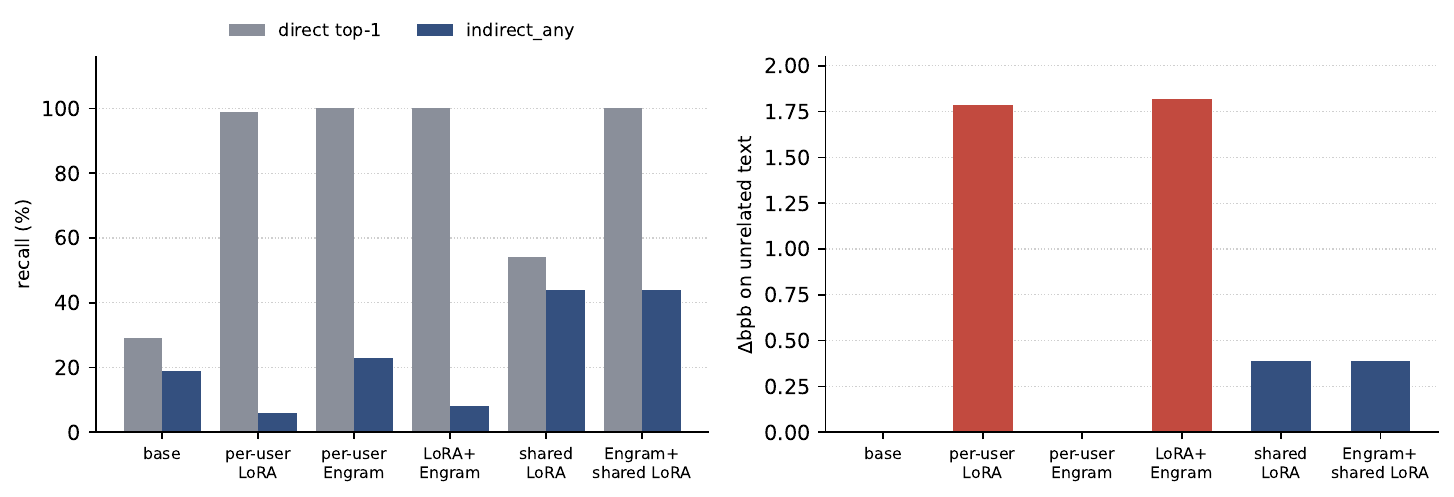}
  \caption{The six parametric conditions on $n{=}20$ test users
  (Mini-Engram-d20, seed S0; bars left-to-right as listed in the
  text). \textbf{Left:} the layered design
  (Engram + shared LoRA) matches per-user LoRA's direct recall
  (100\% vs.\ 99\%)
  while delivering $7.4\times$ its indirect\_any on this seed (44\% vs.\ 6\%);
  neither the per-user Engram (no skill) nor the shared LoRA
  alone (never saw the user's facts) suffices.
  \textbf{Right:} the layered design adds \emph{zero}
  contamination on top of the shared skill ($\Delta$bpb $+0.386$,
  equal to the shared LoRA alone), while per-user LoRA
  and the naive LoRA+Engram stack pay $+1.78$/$+1.82$.
  The retrieval conditions appear in Figure~\ref{fig:pareto-rag}; 3
  seeds in Appendix~\ref{app:layered-seeds} (3-seed mean: layered
  41\%, per-user LoRA 7\%).}
  \label{fig:layered-conditions}
\end{figure}

We compare six parametric conditions, shown as the six bars (in this
order) in Figure~\ref{fig:layered-conditions}:
\textbf{the untouched base},
\textbf{per-user LoRA},
\textbf{the per-user Engram alone} (content, no shared skill),
\textbf{the naive LoRA$+$Engram stack},
\textbf{the shared LoRA alone} (skill, no per-user content), and
\textbf{the layered design} (per-user Engram $+$ shared LoRA);
the retrieval variants are introduced in
Section~\ref{sec:layered-rag}.
We evaluate every per-user combination on $n{=}20$ test users
($n{=}10$ training users held out for the shared LoRA), measuring direct
top-1/top-5 recall on the user's facts (completion-format prompts),
indirect top-1 and indirect\_any on completion-format indirect probes
(``\texttt{Q: <q>\textbackslash nA:}'', greedy 16-token continuation),
and val\_bpb on a held-out 262\,K-token ClimbMix shard
(Figure~\ref{fig:layered-conditions}).

\subsection{Results: the layered design wins on every measure}

The headline is the layered-design-vs-per-user-LoRA contrast
(Figure~\ref{fig:layered-conditions}).
The layered design matches per-user LoRA on direct recall
(100\% vs.\ 99\%) but answers indirect questions
\textbf{$7.4\times$ more often} on this canonical seed (44\% vs.\ 6\%;
$5.6\times$ averaged over three seeds), with
\textbf{$4.6\times$ less damage to unrelated text}---all of it from the
shared skill, none from the per-user Engram---and \textbf{not a
single user out of 20 made worse than the untouched model} (against
17 of 20 for the LoRA), all at 88\,KB per user instead of 14.2\,MB.
The gap is stable across three seeds (paired-bootstrap 95\% CI on the
layered$-$LoRA indirect difference $[+31,+37]$\,pp; per-seed spread in
Appendix~\ref{app:layered-seeds}).

H1 holds: bolting an Engram onto a per-user LoRA does not rescue its
reasoning---the LoRA's damage dominates the combination. H2 holds on
all three counts: adding the per-user Engram on top of the shared
skill costs \emph{zero} extra damage to unrelated text (its
$\Delta$bpb equals the shared LoRA's alone), keeps direct recall at
100\%, and roughly doubles indirect accuracy over either the Engram
alone or the untouched model. Neither piece works on its own---the
Engram has the facts but no reasoning, the shared skill has reasoning
but has never seen the user's facts---and only together do they win.

\subsection{Does RAG just close the gap instead?}
\label{sec:layered-rag}
\label{sec:qwen-rag}

The obvious objection to the layered design is: why not just
retrieve the facts and put them in the prompt? We test this two ways
(the naive-RAG and RAG+shared-LoRA conditions in the context-cost
plot, Appendix~\ref{app:qwen-rag-prompt}, Figure~\ref{fig:pareto-rag}). On
the Engram base itself, putting facts in the prompt sits
\emph{below} the layered design---the base LM never learned to use
an in-context fact list, and even
handing it the exact right facts does not help, because the missing
piece is the reasoning skill, not the facts. The fairer test feeds the
retrieved facts to a real instruction-tuned model
(Qwen2.5-3B-Instruct, about $2.5\times$ larger). Even there, the
layered design---with no retrieval and an empty prompt---lands within
a few points of the larger model, and adding retrieval on top of the
shared skill passes it once context is free. The plain rule: no
room in the prompt $\to$ use the layered design; room to spare $\to$
add retrieval on top of the shared skill.

\subsection{RAG-vs-Engram trade-offs sharpen as the KB grows}
\label{sec:rag-scale}

The experiments above used each user's native 34 facts. Production fact counts are
typically larger ($10^2$--$10^3$: chat history, calendar,
preferences, contacts). We augment each test user's 34 facts
with distractors sampled from the 30-user schema-family pool
(\texttt{u000--u029} minus the active user; 1008 facts) to reach
$N \in \{34, 100, 200, 300, 500, 1000\}$. Indirect probes are
unchanged; the retriever now discriminates the gold fact among
$N$ candidates. The layered design is unaffected: the per-user
Engram table holds only the test user's facts, independent of
population size.

\begin{figure}[t]
  \centering
  \includegraphics[width=\linewidth]{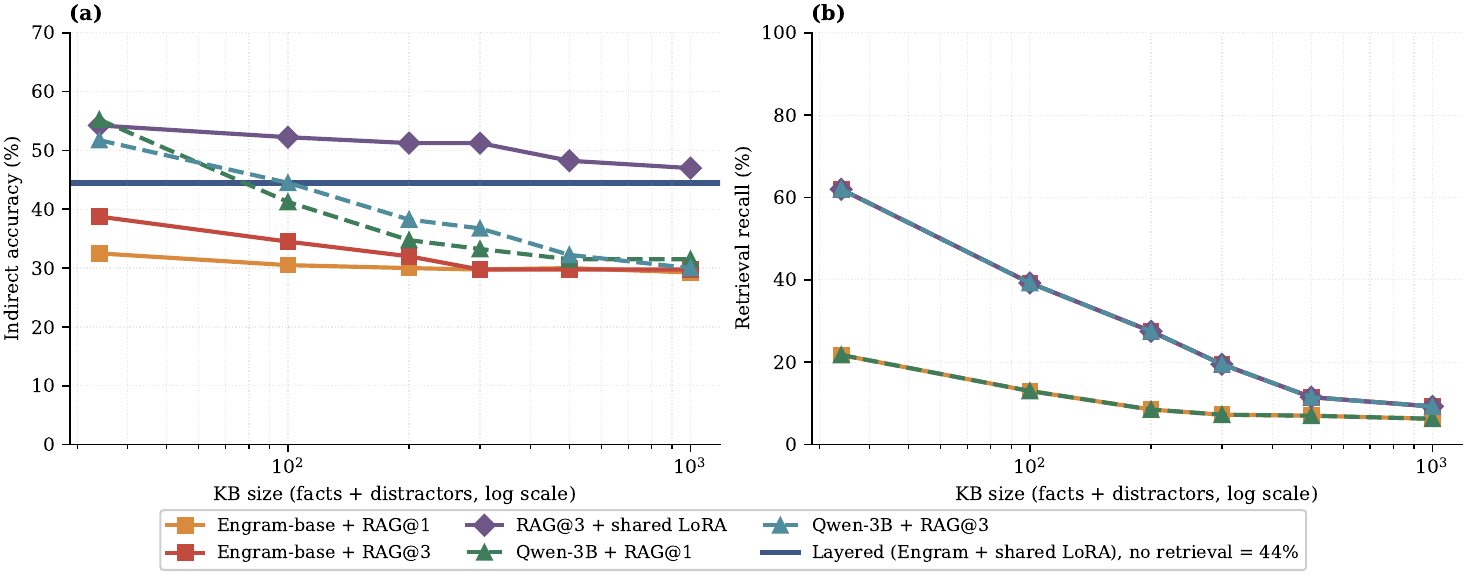}
  \caption{Indirect-reasoning accuracy and retrieval recall vs.\ KB size
  ($n{=}20$ users, 20 indirect probes each; KB = test user's 34 facts +
  distractors sampled from the 30-user schema-family pool, log $x$-axis,
  $N \in \{34, 100, 200, 300, 500, 1000\}$). \textbf{(a)} the layered
  design (horizontal solid line at 44\%) is invariant to KB size; the
  retrieval-based conditions degrade monotonically. \textbf{Naive RAG
  and Qwen-3B + RAG fall below the layered design at KB $\geq$ 100; at
  KB=1000 the gap is
  14\,pp.} RAG top-3 + shared LoRA is the most robust to KB growth
  (54\% $\to$ 47\%) because the shared reasoning skill compensates when
  retrieval misses, though its lead over the layered design shrinks from 10\,pp to 3\,pp
  across the sweep. \textbf{(b)} Retrieval recall (fraction of probes
  where the retrieved set covers every fact the question needs)
  drops by 7$\times$ for top-3 (62\% $\to$ 9\%) between KB=34 and
  KB=1000, which is the mechanistic driver of (a).}
  \label{fig:rag-scale}
\end{figure}


Retrieval degrades quickly: as the candidate pool grows,
top-3 recall on \texttt{all-MiniLM-L6-v2} collapses several-fold
purely from more candidates in the same embedding space, which
drives the accuracy drop. Naive RAG, and even Qwen-3B + RAG, fall
below the layered design once the KB passes $\sim$100 facts, while
the layered design holds
flat---its per-user table never grows. The clean reading
(Figure~\ref{fig:rag-scale}): \emph{which method wins shifts with
deployment scale}. At a few dozen facts, RAG on a larger model is
the accuracy ceiling and the layered design is the zero-context
option; by a few hundred facts, the layered design decisively wins.

\paragraph{Where each side's failure comes from.} The two
limits are mechanistically distinct. Engram's density ceiling
(Section~\ref{sec:density}) comes from forward-pass interference
among co-active rows inside one user's table, and is bounded by
per-user fact count. RAG's retrieval ceiling comes from
nearest-neighbor confusion across an ever-growing candidate
pool, and is bounded by whatever pool the retriever sees
(per-user, per-tenant, or per-corpus). For $10^2$--$10^3$
facts/user, Engram density is the limiting factor on the
parametric side; for $10^4$+ facts in a shared corpus,
retrieval is the limiting factor on the RAG side. The two
costs scale on different axes and cross around $N \approx 100$.

\section{Multi-tenant serving system}
\label{sec:serving}

\begin{figure}[t]
  \centering
  \includegraphics[width=0.74\linewidth]{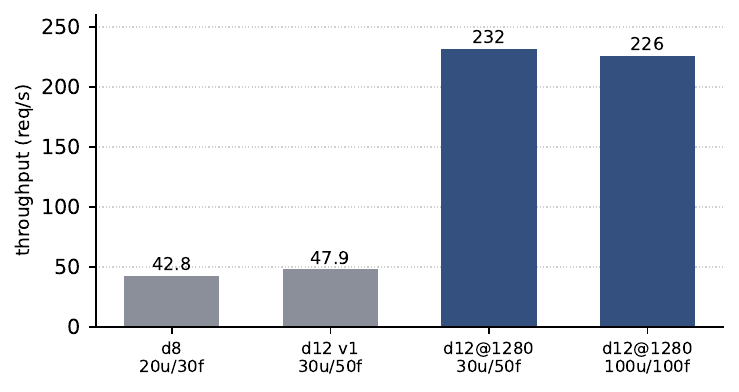}
  \caption{EngramServer throughput across four deployment scales
  on one idle Blackwell GPU. On the ablation-optimal d12@1280 the
  server reaches 232\,req/s (30\,u/50\,f) and holds 226\,req/s at
  100\,u/100\,f---within 3\% as tenants triple---at 4.4\,ms p50
  latency, and a sub-millisecond override apply (0.03\,ms p50; the
  larger per-component figures in Appendix~\ref{app:cdf} are from an
  earlier, slower benchmark on the smaller \texttt{d12}~v2 checkpoint),
  88\,KB/user, and 62\%/96\% top-1/top-5 own-fact recall.}
  \label{fig:serving-scale}
\end{figure}

We implement \texttt{EngramServer}, a $\sim$50-line wrapper
around an Engram-pretrained model. The base and global tables stay
on the GPU; each user's small override map lives in CPU memory. A
request resolves the user, writes their rows, runs the forward pass,
and restores the originals---no router, no custom kernel, no graph
rewrite. Writing and restoring the rows are cheap relative to the
forward pass in every configuration we measured---a sub-millisecond
array write on the ablation-optimal \texttt{d12@1280}
(Figure~\ref{fig:serving-scale}), and at most $\approx$2.2\,ms each in
the earlier per-component breakdown on the smaller \texttt{d12}~v2
checkpoint (Appendix~\ref{app:cdf})---and the frozen forward pass
dominates the rest.

\paragraph{It scales flat in the number of users.}
Figure~\ref{fig:serving-scale} reports throughput, latency,
storage, and recall across four deployment scales.

Per-request work does not depend on how many users share the server,
so throughput and latency barely move as users triple
(Figure~\ref{fig:serving-scale}); each extra user costs one small
row-swap and no shared state. Two users are never in the table at the
same time, so one cannot read another's facts: cross-user leakage is
zero by construction. (The only residual overlap is coincidence---two
users with the same favorite spice---and it is small.) By contrast,
serving per-user LoRAs needs custom CUDA kernels and batch-routing
(S-LoRA~\citep{sheng2024slora}, Punica~\citep{chen2024punica}); here
the write is just an array assignment. Batched multi-user serving
would add one gather per Engram layer, which we do not benchmark.

\section{Discussion and limitations}
\label{sec:discussion}

\subsection{When to use what}
\label{sec:when-to-use}

Which method to use turns on three things: \emph{how many facts
each user has}, \emph{whether the queries need indirect
reasoning}, and \emph{whether context budget or per-user storage
is the tighter constraint}. Table~\ref{tab:when-to-use} gives the
recommended method for each case.

\begin{table}[h]
  \centering
  \caption{Cheat sheet for which method to use.
  ``invariant'' = the number does not
  degrade with KB / population size.}
  \label{tab:when-to-use}
  \resizebox{\linewidth}{!}{
  \begin{tabular}{llll}
    \toprule
    case & recommended method & per-user cost & headline metric \\
    \midrule
    $<$10 facts/user, direct recall only & ICL / strong-retriever RAG & 0\,KB persistent & near-perfect recall \\
    10--1000 facts/user, indirect matters & layered design (Engram + shared LoRA) & \textbf{88\,KB}, 0 ctx toks & 44\% indirect (invariant in KB) \\
    10--1000 facts/user, ctx budget free & RAG top-3 + shared LoRA & 88\,KB + 44 ctx toks & 54\% indirect (KB=34) \\
    enterprise, 100\% direct top-1, $<$10K users & per-user LoRA (S-LoRA serving) & 14.2\,MB & 100\% direct, indirect cost \\
    long-form / open-domain / shared corpus & RAG & 0\,KB persistent & wins LOCOMO open-domain $+53$\% \\
    \bottomrule
  \end{tabular}}
\end{table}

The most important row is the second: the layered design is the right
choice for the common case (10--1000 facts/user, indirect
reasoning matters, many users), and decisively beats RAG with a
$\sim$2.5$\times$ larger instruction-tuned LM at $N \geq 100$.
\emph{Any} method that reaches 100\% direct top-1 pays a price in
indirect reasoning; the question is whether that price is baked
in (a global LoRA), manageable (a local Engram, where the trigger
N-gram is the contract), or set by retrieval (RAG, where the
encoder must pick the gold fact out of $N$ candidates).

\subsection{Open limitations}
\label{sec:limitations}

\paragraph{Shared multi-hop reasoning gap.}
Both LoRA and Engram are surface-trigger keyed; neither
composes facts across triggers (``if my doctor is Patel and
Patel works at Globex, what is my doctor's employer?'').
User-as-Engram inherits this gap from per-user LoRA. A candidate
fix is to mix recite-then-reason traces into Engram pretraining;
we leave it to future work.

\paragraph{Within-user density ceiling.}
Joint OPT at 1000 facts reaches 35\% top-1. The slot space is
highly sparse (${<}\,1\%$ occupancy), so the constraint is
forward-pass interference among co-active rows, not hash
collisions. Per-user scoped tables (load only the relevant rows
for the current query, via a small classifier) is a candidate
future fix; another is the multi-fact-in-the-loss recipe change
in Section~\ref{sec:future}.

\paragraph{Engram pretraining required.}
Our method assumes an Engram-pretrained base. We trained four
ourselves at 178\,M, 339\,M, 625\,M, and 1.22\,B total
parameters; production deployment depends on either DeepSeek
releasing weights or pretraining your own ($\sim$10\,h on a
single Blackwell GPU at our 625\,M scale; production-scale
Engrams require correspondingly larger compute budgets).

\subsection{Future work: training-recipe changes}
\label{sec:future}

Three changes to the training recipe target the limits we
documented, and we expect more from them per unit of compute than
from simply making the model bigger. \emph{(1) Put many facts in the
loss during training}, so the model learns to hold several at once.
We tested a lighter version of this by finetuning d12@1280, and it
helps at high density---but only by reaching the same recall in fewer
optimization steps, not by raising the ceiling
(Appendix~\ref{sec:mf-finetune}, Figure~\ref{fig:mf-finetune}). This
pins the density ceiling on inference-time optimization, not
pretraining. \emph{(2) Train with random per-user offsets to the
hash}, which would let the gate read rows it never saw in training and
bring back a per-user-salt option. \emph{(3) Mix in ``recall the fact,
then reason'' examples}, the most direct attack on the chaining gap.

\paragraph{A negative result worth recording.} We tried improving the
shared skill by training it on reasoning traces from a teacher model.
It cut the contamination but also \emph{lowered} accuracy
(Appendix~\ref{app:extended}). The reason is instructive: those traces
assume the model first writes out a thinking step, while our test
reads only the final answer. The recipe and the way you score it have
to be designed together---you cannot mix and match them. Remaining
layered follow-ups: a larger shared-skill training set, training the
per-user table and the shared skill together, and starting from a
chat-tuned base.

\section{Related Work}
\label{sec:related-work}

\paragraph{Memory architectures and adjacent work.}
Engram~\citep{cheng2026engram} sits in a long line of trainable
key--value memory modules~\citep{geva2021kv,lample2019pkm,he2024peer,berges2025memory,ultramem2025,overencoding2025,scone2025,pagnoni2025blt,superbpe2025},
but none of them addresses writing a single user's fact in at
inference time. Stacking edits without a combiner echoes LoRA
stacking for Stable Diffusion~\citep{lora_diffusion}
(LoRAHub~\citep{huang2023lorahub} instead trains one). A mirror-image
line keeps memory \emph{outside} the weights but executable---User as
Code~\citep{li2026userascode} compiles a user's history into typed
state and rules---which we contrast with our in-weights edits in
Section~\ref{sec:experiments}.

\paragraph{Memory systems for LLM agents.}
A large body of work gives an agent long-term memory by keeping
facts \emph{outside} the weights and retrieving them at query
time: paging systems~\citep{packer2023memgpt}; extraction- and
graph-based stores~\citep{chhikara2025mem0,xu2025amem,rasmussen2025zep,li2025memos,hu2026evermemos,wang2026memmachine};
and a line that learns \emph{how} to read and write the
store~\citep{yan2025memoryr1,yu2026agemem,wang2025memalpha}, with
surveys cataloguing the
space~\citep{zhang2024memorysurvey,wu2025humantoai,du2026memorysurvey}.
None writes a fact \emph{into} the model. (One concurrent system
even shares our name~\citep{patel2025engram} while doing the
opposite---it orchestrates memories in natural language.) We use
Mem0- and MemMachine-style retrieval as our strongest baselines
(Section~\ref{sec:memory-systems}) and ask a different question:
\emph{when} does writing a fact into the weights beat keeping it in
a retrievable store~\citep{pollertlam2026beyondcontext}?

\paragraph{Personalizing language models.} Adapting a model to an
individual user is an active area in its own
right~\citep{zhang2024personalization,liu2025personalizedllm,xu2026personalizedagents},
with benchmarks such as LaMP~\citep{salemi2024lamp} measuring
personalized generation. Most methods personalize through the
prompt or a retrieval store; the per-user LoRA line above is the
parametric alternative, and User as Engram is our attempt to win
the locality of a prompt at the zero-context cost of a weight
edit.

\paragraph{Retrieval and non-parametric memory.}
Retrieval-augmented generation~\citep{lewis2020rag,gao2023ragsurvey}
is the dominant non-parametric route, spanning retrieve-then-read
pretraining, large-scale retrieval pretraining, and self-reflective
retrieval~\citep{guu2020realm,khandelwal2020knnlm,borgeaud2022retro,izacard2023atlas,asai2024selfrag}.
Its quality is bounded by the encoder's ability to surface the right
evidence among many candidates---exactly the failure mode we measure
as the knowledge base grows (Section~\ref{sec:rag-scale}).

\paragraph{Personal memory benchmarks.} Long-term personal memory
is evaluated by LOCOMO~\citep{maharana2024locomo},
LongMemEval~\citep{wu2024longmemeval},
BEAM~\citep{liu2025beam} (multi-turn evolving beliefs), and
PersonaMem-v2~\citep{personamem2025}. We use synthetic per-user
fact corpora for the within-user fact-density sweeps (tighter
control over the parameter we vary), and LOCOMO
(Section~\ref{sec:locomo}) for end-to-end evaluation against
retrieval baselines. We chose LOCOMO over LongMemEval and BEAM
because its per-question evidence pointers let us train a
per-fact Engram row and a retrieval entry against the same gold
sentence---an apples-to-apples ``same gold, different
method'' comparison. LongMemEval and BEAM remain open for
future work.

\paragraph{Knowledge editing and recitation.}
Knowledge editing changes a fact the model already
holds---locating and rewriting weights~\citep{meng2022rome,meng2023memit,dai2022knowledge,yao2023editing},
or routing around them~\citep{mitchell2022mend,mitchell2022serac,wang2024wise}---with
ripple-effect benchmarks measuring the
fallout~\citep{mquake2023,rippleedits2023,counterfact2022}. We do the
opposite: we \emph{add} facts the model never saw and want everything
else left alone. Recitation methods, which surface a stored fact
before reasoning over it~\citep{srrag2024,sun2023recitation}, we reuse
in Section~\ref{sec:negative}.

\paragraph{Complementary learning systems and continual learning.}
The extra loss a per-user LoRA imposes on unrelated text is a modern
instance of \emph{catastrophic interference}: training on
something new overwrites what was there
before~\citep{mccloskey1989catastrophic}, which decades of
continual-learning work fight by protecting important
weights~\citep{kirkpatrick2017ewc} or replaying old data, with
the problem persisting for large language
models~\citep{shi2024continual}. The brain avoids it with a
different architecture, and the word \emph{engram} points to how:
an engram is the physical trace a memory leaves in neural
tissue~\citep{semon1921mneme}, and \emph{engram cells}---sparse
neuron populations whose reactivation triggers recall---sit
largely in the hippocampus~\citep{tonegawa2015engram}, while
general skills live in the slow, distributed neocortex.
Complementary learning systems
theory~\citep{mcclelland1995cls,kumaran2016cls} argues this split
is what lets a new episode be written without overwriting old
skills. User as Engram is the same division made architectural:
the per-user Engram row is the fast, sparse, local trace (the
hippocampal engram); the shared LoRA and frozen backbone are the
slow, distributed skill (the neocortex); and our measured locality
($\Delta$bpb $+0.00005$ on unrelated text,
Section~\ref{sec:negative}) is the engineering form of the
\emph{pattern separation} that keeps the two from interfering. The
analogy is also honest about where our mechanism falls short: the
hippocampus performs \emph{pattern completion}---recovering a whole
memory from a partial or indirect cue---whereas our gated
trigger-N-gram lookup fires only on a near-exact surface match, which
is exactly the multi-hop and paraphrase gap of
Sections~\ref{sec:multihop} and~\ref{sec:paraphrase}. The biology
thus predicts the precise capability we still lack. Where
continual learning asks one set of weights to absorb new facts
gracefully, we sidestep the interference by writing each fact to
its own address and leaving the shared weights untouched.

\section{Conclusion}
\label{sec:conclusion}

ersonal memory is two jobs, not one: remembering a
user's specific facts, and the general skill of reasoning with
them. The standard in-weights recipe asks one per-user adapter to
do both, and that is the root of the trouble: an adapter that must
\emph{hold} the facts cannot help also reshaping how the model
thinks, so remembering and reasoning end up pulling on the same
weights and the user pays for it in damaged knowledge and weaker
reasoning. User as Engram separates the two---a user's facts
become a small, local edit, and the reasoning skill lives in one
model everyone shares.

The deeper lesson is that the trouble is not unique to adapters.
Every method that reaches near-perfect recall pays the same price
in weaker reasoning over the stored facts; what differs is only
\emph{where} the price lands. A per-user LoRA pays it in the shared
weights, on every query and stacking with every user; retrieval
pays it in a search index that degrades as the candidate pool
grows; User as Engram pays it at the one address its fact occupies,
so the cost grows with a single user's facts and nothing else.

A century ago Richard Semon coined \emph{engram} for the trace
an experience etches into living tissue---a mark local enough
that one memory does not erase the next. The brain pairs that
sparse, local trace with a slow cortex that learns how to use
it, and the pairing is what lets a person remember a new fact
without forgetting how to think. Maya's assistant needs the same
discipline. When we stop asking one set of weights to be both
the memory and the mind, and instead give each user a small
private trace beside a skill everyone shares, personalization
stops fighting the model it runs on. That, more than any single
number, is what \emph{User as Engram} is for.

\section*{Acknowledgements}

Pine Copilot, Claude Code, and Claude Opus~4.8 were used during
this research. We thank BSQL Networking for hosting the NVIDIA
RTX Pro~6000 GPU on which these experiments were run.

\bibliographystyle{plainnat}

\begin{thebibliography}{50}

\bibitem[Li(2026)]{li2026userascode}
B.~Li.
\newblock User as code: Executable memory for personalized agents.
\newblock arXiv:2606.16707, 2026.

\bibitem[Semon(1921)]{semon1921mneme}
R.~Semon.
\newblock \emph{The Mneme}.
\newblock George Allen \& Unwin, London, 1921.
\newblock (English translation of \emph{Die Mneme}, 1904; origin of
  the term ``engram'').

\bibitem[McClelland et~al.(1995)]{mcclelland1995cls}
J.~L. McClelland, B.~L. McNaughton, and R.~C. O'Reilly.
\newblock Why there are complementary learning systems in the
  hippocampus and neocortex: Insights from the successes and failures
  of connectionist models of learning and memory.
\newblock \emph{Psychological Review}, 102(3):419--457, 1995.

\bibitem[Tonegawa et~al.(2015)]{tonegawa2015engram}
S.~Tonegawa, X.~Liu, S.~Ramirez, and R.~Redondo.
\newblock Memory engram cells have come of age.
\newblock \emph{Neuron}, 87(5):918--931, 2015.

\bibitem[Kumaran et~al.(2016)]{kumaran2016cls}
D.~Kumaran, D.~Hassabis, and J.~L. McClelland.
\newblock What learning systems do intelligent agents need?
  Complementary learning systems theory updated.
\newblock \emph{Trends in Cognitive Sciences}, 20(7):512--534, 2016.

\bibitem[Cheng et~al.(2026)]{cheng2026engram}
X.~Cheng, W.~Zeng, D.~Dai, Q.~Chen, B.~Wang, Z.~Xie, K.~Huang, X.~Yu, Z.~Hao,
  Y.~Li, H.~Zhang, H.~Zhang, D.~Zhao, and W.~Liang.
\newblock Conditional memory via scalable lookup: A new axis of sparsity for
  large language models.
\newblock arXiv:2601.07372, January 2026.

\bibitem[Vaswani et~al.(2017)]{vaswani2017transformer}
A.~Vaswani, N.~Shazeer, N.~Parmar, J.~Uszkoreit, L.~Jones, A.~N. Gomez,
  L.~Kaiser, and I.~Polosukhin.
\newblock Attention is all you need.
\newblock NeurIPS 2017.

\bibitem[Brown et~al.(2020)]{brown2020gpt3}
T.~Brown et~al.
\newblock Language models are few-shot learners.
\newblock NeurIPS 2020.

\bibitem[Radford et~al.(2019)]{radford2019gpt2}
A.~Radford, J.~Wu, R.~Child, D.~Luan, D.~Amodei, and I.~Sutskever.
\newblock Language models are unsupervised multitask learners ({GPT-2}).
\newblock OpenAI technical report, 2019.

\bibitem[Min et~al.(2022)]{min2022rethinking}
S.~Min, X.~Lyu, A.~Holtzman, M.~Artetxe, M.~Lewis, H.~Hajishirzi, and
  L.~Zettlemoyer.
\newblock Rethinking the role of demonstrations: What makes in-context learning
  work?
\newblock EMNLP 2022.

\bibitem[Lewis et~al.(2020)]{lewis2020rag}
P.~Lewis, E.~Perez, A.~Piktus, F.~Petroni, V.~Karpukhin, N.~Goyal,
  H.~K\"uttler, M.~Lewis, W.~Yih, T.~Rockt\"aschel, S.~Riedel, and D.~Kiela.
\newblock Retrieval-augmented generation for knowledge-intensive {NLP} tasks.
\newblock NeurIPS 2020.

\bibitem[Gao et~al.(2023)]{gao2023ragsurvey}
Y.~Gao et~al.
\newblock Retrieval-augmented generation for large language models: A survey.
\newblock arXiv:2312.10997, 2023.

\bibitem[Shazeer et~al.(2017)]{shazeer2017moe}
N.~Shazeer, A.~Mirhoseini, K.~Maziarz, A.~Davis, Q.~Le, G.~Hinton, and J.~Dean.
\newblock Outrageously large neural networks: The sparsely-gated
  mixture-of-experts layer.
\newblock ICLR 2017.

\bibitem[Dai et~al.(2024)]{dai2024deepseekmoe}
D.~Dai et~al.
\newblock {DeepSeekMoE}: Towards ultimate expert specialization in
  mixture-of-experts language models.
\newblock arXiv:2401.06066, 2024.

\bibitem[Mangrulkar et~al.(2022)]{mangrulkar2022peft}
S.~Mangrulkar et~al.
\newblock {PEFT}: State-of-the-art parameter-efficient fine-tuning methods.
\newblock \url{https://github.com/huggingface/peft}, 2022.

\bibitem[Jordan et~al.(2024)]{jordan2024muon}
K.~Jordan et~al.
\newblock {Muon}: An optimiser for hidden layers in neural networks.
\newblock GitHub / blog post, 2024.

\bibitem[Huang et~al.(2023)]{huang2023lorahub}
C.~Huang et~al.
\newblock {LoRAHub}: Efficient cross-task generalization via dynamic {LoRA}
  composition.
\newblock arXiv:2307.13269, 2023.

\bibitem[Wu et~al.(2024)]{wu2024longmemeval}
D.~Wu et~al.
\newblock {LongMemEval}: Benchmarking chat assistants on long-term memory.
\newblock arXiv 2024.

\bibitem[Maharana et~al.(2024)]{maharana2024locomo}
A.~Maharana et~al.
\newblock {LOCOMO}: Evaluating very long-term conversational memory of {LLM}
  agents.
\newblock ACL 2024.

\bibitem[Tavakoli et~al.(2025)]{liu2025beam}
M.~Tavakoli, A.~Salemi, C.~Ye, M.~Abdalla, H.~Zamani, and J.~R. Mitchell.
\newblock Beyond a million tokens: Benchmarking and enhancing long-term memory
  in {LLM}s ({BEAM}).
\newblock arXiv:2510.27246, 2025.

\bibitem[Jiang et~al.(2025)]{personamem2025}
B.~Jiang, Y.~Yuan, M.~Shen, Z.~Hao, Z.~Xu, Z.~Chen, Z.~Liu, A.~R. Vijjini,
  J.~He, H.~Yu, R.~Poovendran, G.~Wornell, L.~Ungar, D.~Roth, S.~Chen, and
  C.~J. Taylor.
\newblock {PersonaMem-v2}: Towards personalized intelligence via learning
  implicit user personas and agentic memory.
\newblock arXiv:2512.06688, 2025.

\bibitem[Karpathy(2026)]{karpathy2026nanochat}
A.~Karpathy.
\newblock nanochat: an experimental training harness for LLMs.
\newblock \url{https://github.com/karpathy/nanochat}, 2026.

\bibitem[Hu et~al.(2022)]{hu2022lora}
E.~Hu, Y.~Shen, P.~Wallis, Z.~Allen-Zhu, Y.~Li, S.~Wang, L.~Wang, and W.~Chen.
\newblock {LoRA}: Low-rank adaptation of large language models.
\newblock ICLR 2022.

\bibitem[Houlsby et~al.(2019)]{houlsby2019adapters}
N.~Houlsby et~al.
\newblock Parameter-efficient transfer learning for {NLP}.
\newblock ICML 2019.

\bibitem[Su et~al.(2025)]{su2024prag}
W.~Su et~al.
\newblock Parametric retrieval-augmented generation ({PRAG}).
\newblock arXiv:2501.15915, 2025.

\bibitem[Tan et~al.(2025)]{tan2025dyprag}
Z.~Tan et~al.
\newblock {DyPRAG}: Dynamic parametric {RAG}.
\newblock arXiv:2505.19386, 2025.

\bibitem[Chen et~al.(2025)]{distilledprag2025}
J.~Chen, H.~Zhang, L.~Pang, Y.~Tong, H.~Zhou, Y.~Zhan, W.~Lin, and Z.~Zheng.
\newblock Privacy-preserving reasoning with knowledge-distilled parametric
  retrieval-augmented generation ({DistilledPRAG}).
\newblock arXiv:2509.01088, 2025.

\bibitem[Tan et~al.(2024)]{tan2024oppu}
Z.~Tan, Q.~Liu, and M.~Jiang.
\newblock Democratizing large language models via personalized
  parameter-efficient fine-tuning ({OPPU}).
\newblock EMNLP 2024 (arXiv:2402.04401).

\bibitem[Zhuang et~al.(2024)]{zhuang2024hydra}
Y.~Zhuang et~al.
\newblock {HYDRA}: Per-user adapters for personalised {LLMs}.
\newblock arXiv 2024.

\bibitem[Bini et~al.(2025)]{memlora2025}
M.~Bini, O.~Bohdal, U.~Michieli, Z.~Akata, M.~Ozay, and T.~Ceritli.
\newblock {MemLoRA}: Distilling expert adapters for on-device memory systems.
\newblock arXiv:2512.04763, 2025.

\bibitem[Charakorn et~al.(2025)]{t2l2024}
R.~Charakorn, E.~Cetin, Y.~Tang, and R.~T. Lange.
\newblock Text-to-{LoRA}: Instant transformer adaption.
\newblock arXiv:2506.06105, 2025.

\bibitem[Tan et~al.(2024b)]{tan2024perpcs}
Z.~Tan et~al.
\newblock {PER-PCS}: Per-user post-hoc {LoRA} composition.
\newblock arXiv 2024.

\bibitem[Sheng et~al.(2024)]{sheng2024slora}
Y.~Sheng, S.~Cao, D.~Li, et~al.
\newblock {S-LoRA}: Serving thousands of concurrent {LoRA} adapters.
\newblock arXiv:2311.03285, 2024.

\bibitem[Chen et~al.(2024)]{chen2024punica}
L.~Chen, Z.~Ye, Y.~Wu, et~al.
\newblock Punica: Multi-tenant {LoRA} serving.
\newblock MLSys 2024.

\bibitem[Lample et~al.(2019)]{lample2019pkm}
G.~Lample, A.~Sablayrolles, M.~Ranzato, L.~Denoyer, and H.~J\'egou.
\newblock Large memory layers with product keys.
\newblock NeurIPS 2019.

\bibitem[He(2024)]{he2024peer}
P.~He.
\newblock {PEER}: Mixture of one million experts.
\newblock arXiv 2024.

\bibitem[Berges et~al.(2025)]{berges2025memory}
V.~Berges, B.~O\u{g}uz, D.~Haziza, W.~Yih, L.~Zettlemoyer, and G.~Ghosh.
\newblock Memory layers at scale.
\newblock ICML 2025.

\bibitem[Huang et~al.(2024b)]{ultramem2025}
Z.~Huang, Q.~Min, H.~Huang, D.~Zhu, Y.~Zeng, R.~Guo, and X.~Zhou.
\newblock Ultra-sparse memory network ({Ultra-Mem}).
\newblock arXiv:2411.12364, 2024 (ICLR 2025).

\bibitem[Huang et~al.(2025)]{overencoding2025}
J.~Huang et~al.
\newblock {OverEncoding}: hashed N-gram embeddings via averaging.
\newblock 2025.

\bibitem[Yu et~al.(2025)]{scone2025}
L.~Yu et~al.
\newblock {SCONE}: scalable contextual N-gram embeddings.
\newblock 2025.

\bibitem[Pagnoni et~al.(2025)]{pagnoni2025blt}
A.~Pagnoni, R.~Pasunuru, P.~Rodriguez, et~al.
\newblock {BLT}: byte latent transformer with hashed N-gram embeddings.
\newblock arXiv:2412.09871, 2025.

\bibitem[Liu et~al.(2025)]{superbpe2025}
A.~Liu et~al.
\newblock {SuperBPE}: word-level BPE for compositional patterns.
\newblock 2025.

\bibitem[CivitAI(2024)]{lora_diffusion}
{CivitAI Community}.
\newblock {LoRA} stacking patterns for Stable Diffusion.
\newblock \url{https://civitai.com/}, 2024.

\bibitem[Meng et~al.(2022)]{meng2022rome}
K.~Meng, D.~Bau, A.~Andonian, and Y.~Belinkov.
\newblock Locating and editing factual associations in {GPT} ({ROME}).
\newblock NeurIPS 2022.

\bibitem[Meng et~al.(2023)]{meng2023memit}
K.~Meng et~al.
\newblock {MEMIT}: Mass-editing memory in a transformer.
\newblock ICLR 2023.

\bibitem[Cohen et~al.(2023)]{mquake2023}
R.~Cohen et~al.
\newblock Evaluating the ripple effects of knowledge editing in language models
  ({MQuAKE}).
\newblock 2023.

\bibitem[Cohen et~al.(2024)]{rippleedits2023}
R.~Cohen et~al.
\newblock {RippleEdits}: A benchmark for ripple effects of model editing.
\newblock 2024.

\bibitem[Meng et~al.(2022b)]{counterfact2022}
K.~Meng et~al.
\newblock {CounterFact}: a counterfactual editing benchmark.
\newblock 2022.

\bibitem[Wu et~al.(2025)]{srrag2024}
D.~Wu, J.-C. Gu, K.-W. Chang, and N.~Peng.
\newblock Self-routing {RAG}: Binding selective retrieval with knowledge
  verbalization.
\newblock arXiv:2504.01018, 2025.

\bibitem[Sun et~al.(2023)]{sun2023recitation}
Z.~Sun et~al.
\newblock Recitation-augmented language models.
\newblock ICLR 2023.

\bibitem[Qwen Team(2025a)]{qwen2025qwen25}
A.~Yang, B.~Yang, B.~Zhang, et~al.
\newblock Qwen2.5 technical report.
\newblock arXiv:2412.15115, 2025.

\bibitem[Qwen Team(2025b)]{qwen2025qwen3}
A.~Yang, A.~Li, B.~Yang, et~al.
\newblock Qwen3 technical report.
\newblock arXiv:2505.09388, 2025.

\bibitem[Grattafiori et~al.(2024)]{grattafiori2024llama3}
A.~Grattafiori, A.~Dubey, A.~Jauhri, et~al.
\newblock The Llama 3 herd of models.
\newblock arXiv:2407.21783, 2024.

\bibitem[Jiang et~al.(2023)]{jiang2023mistral}
A.~Q. Jiang, A.~Sablayrolles, A.~Mensch, et~al.
\newblock Mistral 7B.
\newblock arXiv:2310.06825, 2023.

\bibitem[DeepSeek-AI(2024)]{deepseekai2024v3}
DeepSeek-AI.
\newblock DeepSeek-V3 technical report.
\newblock arXiv:2412.19437, 2024.

\bibitem[Reimers \& Gurevych(2019)]{reimers2019sbert}
N.~Reimers and I.~Gurevych.
\newblock Sentence-BERT: Sentence embeddings using Siamese BERT-networks.
\newblock EMNLP-IJCNLP 2019.

\bibitem[Wang et~al.(2020)]{wang2020minilm}
W.~Wang, F.~Wei, L.~Dong, H.~Bao, N.~Yang, and M.~Zhou.
\newblock MiniLM: Deep self-attention distillation for task-agnostic compression of pre-trained transformers.
\newblock NeurIPS 2020.

\bibitem[Zheng et~al.(2023)]{zheng2023mtbench}
L.~Zheng, W.-L. Chiang, Y.~Sheng, et~al.
\newblock Judging LLM-as-a-judge with MT-Bench and Chatbot Arena.
\newblock NeurIPS 2023 Datasets and Benchmarks. arXiv:2306.05685.

\bibitem[Packer et~al.(2023)]{packer2023memgpt}
C.~Packer, S.~Wooders, K.~Lin, et~al.
\newblock MemGPT: Towards LLMs as operating systems.
\newblock arXiv:2310.08560, 2023.

\bibitem[Chhikara et~al.(2025)]{chhikara2025mem0}
P.~Chhikara, D.~Khant, S.~Aryan, T.~Singh, and D.~Yadav.
\newblock Mem0: Building production-ready AI agents with scalable long-term memory.
\newblock arXiv:2504.19413, 2025.

\bibitem[Xu et~al.(2025)]{xu2025amem}
W.~Xu, Z.~Liang, K.~Mei, H.~Gao, J.~Tan, and Y.~Zhang.
\newblock A-MEM: Agentic memory for LLM agents.
\newblock arXiv:2502.12110, 2025.

\bibitem[Rasmussen et~al.(2025)]{rasmussen2025zep}
P.~Rasmussen, P.~Paliychuk, T.~Beauvais, J.~Ryan, and D.~Chalef.
\newblock Zep: A temporal knowledge graph architecture for agent memory.
\newblock arXiv:2501.13956, 2025.

\bibitem[Li et~al.(2025)]{li2025memos}
Z.~Li, S.~Song, H.~Wang, et~al.
\newblock MemOS: An operating system for memory-augmented generation in large language models.
\newblock arXiv:2505.22101, 2025.

\bibitem[Wang et~al.(2026)]{wang2026memmachine}
S.~Wang, E.~Yu, O.~Love, T.~Zhang, T.~Wong, S.~Scargall, and C.~Fan.
\newblock MemMachine: A ground-truth-preserving memory system for personalized AI agents.
\newblock arXiv:2604.04853, 2026.

\bibitem[Hu et~al.(2026)]{hu2026evermemos}
C.~Hu, X.~Gao, Z.~Zhou, et~al.
\newblock EverMemOS: A self-organizing memory operating system for structured long-horizon reasoning.
\newblock arXiv:2601.02163, 2026.

\bibitem[Patel \& Patel(2025)]{patel2025engram}
D.~Patel and S.~Patel.
\newblock ENGRAM: Effective, lightweight memory orchestration for conversational agents.
\newblock arXiv:2511.12960, 2025.

\bibitem[Yan et~al.(2025)]{yan2025memoryr1}
S.~Yan, X.~Yang, Z.~Huang, et~al.
\newblock Memory-R1: Enhancing large language model agents to manage and utilize memories via reinforcement learning.
\newblock arXiv:2508.19828, 2025.

\bibitem[Yu et~al.(2026)]{yu2026agemem}
Y.~Yu, L.~Yao, Y.~Xie, et~al.
\newblock Agentic memory: Learning unified long-term and short-term memory management for LLM agents.
\newblock arXiv:2601.01885, 2026.

\bibitem[Wang et~al.(2025)]{wang2025memalpha}
Y.~Wang, R.~Takanobu, Z.~Liang, et~al.
\newblock Mem-$\alpha$: Learning memory construction via reinforcement learning.
\newblock arXiv:2509.25911, 2025.

\bibitem[Zhang et~al.(2024)]{zhang2024memorysurvey}
Z.~Zhang, X.~Bo, C.~Ma, et~al.
\newblock A survey on the memory mechanism of large language model based agents.
\newblock arXiv:2404.13501, 2024.

\bibitem[Wu et~al.(2025)]{wu2025humantoai}
Y.~Wu, S.~Liang, C.~Zhang, et~al.
\newblock From human memory to AI memory: A survey on memory mechanisms in the era of LLMs.
\newblock arXiv:2504.15965, 2025.

\bibitem[Du(2026)]{du2026memorysurvey}
P.~Du.
\newblock Memory for autonomous LLM agents: Mechanisms, evaluation, and emerging frontiers.
\newblock arXiv:2603.07670, 2026.

\bibitem[Pollertlam \& Kornsuwannawit(2026)]{pollertlam2026beyondcontext}
N.~Pollertlam and W.~Kornsuwannawit.
\newblock Beyond the context window: A cost-performance analysis of fact-based memory vs.\ long-context LLMs for persistent agents.
\newblock arXiv:2603.04814, 2026.

\bibitem[Salemi et~al.(2024)]{salemi2024lamp}
A.~Salemi, S.~Mysore, M.~Bendersky, and H.~Zamani.
\newblock LaMP: When large language models meet personalization.
\newblock ACL 2024. arXiv:2304.11406.

\bibitem[Zhang et~al.(2024)]{zhang2024personalization}
Z.~Zhang, R.~A. Rossi, B.~Kveton, et~al.
\newblock Personalization of large language models: A survey.
\newblock arXiv:2411.00027, 2024.

\bibitem[Liu et~al.(2025)]{liu2025personalizedllm}
J.~Liu, Z.~Qiu, Z.~Li, et~al.
\newblock A survey of personalized large language models: Progress and future directions.
\newblock arXiv:2502.11528, 2025.

\bibitem[Xu et~al.(2026)]{xu2026personalizedagents}
Y.~Xu, Q.~Chen, Z.~Ma, et~al.
\newblock Toward personalized LLM-powered agents: Foundations, evaluation, and future directions.
\newblock arXiv:2602.22680, 2026.

\bibitem[Mitchell et~al.(2022a)]{mitchell2022mend}
E.~Mitchell, C.~Lin, A.~Bosselut, C.~Finn, and C.~D.~Manning.
\newblock Fast model editing at scale.
\newblock ICLR 2022.

\bibitem[Mitchell et~al.(2022b)]{mitchell2022serac}
E.~Mitchell, C.~Lin, A.~Bosselut, C.~D.~Manning, and C.~Finn.
\newblock Memory-based model editing at scale.
\newblock ICML 2022.

\bibitem[Dai et~al.(2022)]{dai2022knowledge}
D.~Dai, L.~Dong, Y.~Hao, Z.~Sui, B.~Chang, and F.~Wei.
\newblock Knowledge neurons in pretrained transformers.
\newblock ACL 2022.

\bibitem[Allen-Zhu \& Li(2025)]{allenzhu2025physics}
Z.~Allen-Zhu and Y.~Li.
\newblock Physics of language models: Part 3.3, knowledge capacity scaling laws.
\newblock ICML 2025. arXiv:2404.05405.

\bibitem[Li(2026b)]{li2026ikp}
B.~Li.
\newblock Incompressible knowledge probes: Estimating black-box {LLM} parameter
  counts via factual capacity.
\newblock arXiv:2604.24827, 2026.

\bibitem[Wang et~al.(2024)]{wang2024wise}
P.~Wang, Z.~Li, N.~Zhang, et~al.
\newblock WISE: Rethinking the knowledge memory for lifelong model editing of large language models.
\newblock NeurIPS 2024.

\bibitem[Yao et~al.(2023)]{yao2023editing}
Y.~Yao, P.~Wang, B.~Tian, et~al.
\newblock Editing large language models: Problems, methods, and opportunities.
\newblock EMNLP 2023.

\bibitem[McCloskey \& Cohen(1989)]{mccloskey1989catastrophic}
M.~McCloskey and N.~J. Cohen.
\newblock Catastrophic interference in connectionist networks: The sequential learning problem.
\newblock \emph{Psychology of Learning and Motivation}, 24:109--165, 1989.

\bibitem[Kirkpatrick et~al.(2017)]{kirkpatrick2017ewc}
J.~Kirkpatrick, R.~Pascanu, N.~Rabinowitz, et~al.
\newblock Overcoming catastrophic forgetting in neural networks.
\newblock \emph{PNAS}, 114(13):3521--3526, 2017.

\bibitem[Shi et~al.(2024)]{shi2024continual}
H.~Shi, Z.~Xu, H.~Wang, et~al.
\newblock Continual learning of large language models: A comprehensive survey.
\newblock arXiv:2404.16789, 2024.

\bibitem[Guu et~al.(2020)]{guu2020realm}
K.~Guu, K.~Lee, Z.~Tung, P.~Pasupat, and M.-W. Chang.
\newblock REALM: Retrieval-augmented language model pre-training.
\newblock ICML 2020.

\bibitem[Khandelwal et~al.(2020)]{khandelwal2020knnlm}
U.~Khandelwal, O.~Levy, D.~Jurafsky, L.~Zettlemoyer, and M.~Lewis.
\newblock Generalization through memorization: Nearest neighbor language models.
\newblock ICLR 2020.

\bibitem[Borgeaud et~al.(2022)]{borgeaud2022retro}
S.~Borgeaud, A.~Mensch, J.~Hoffmann, et~al.
\newblock Improving language models by retrieving from trillions of tokens.
\newblock ICML 2022.

\bibitem[Izacard et~al.(2023)]{izacard2023atlas}
G.~Izacard, P.~Lewis, M.~Lomeli, et~al.
\newblock Atlas: Few-shot learning with retrieval augmented language models.
\newblock \emph{JMLR}, 24(251):1--43, 2023.

\bibitem[Asai et~al.(2024)]{asai2024selfrag}
A.~Asai, Z.~Wu, Y.~Wang, A.~Sil, and H.~Hajishirzi.
\newblock Self-RAG: Learning to retrieve, generate, and critique through self-reflection.
\newblock ICLR 2024.

\bibitem[Geva et~al.(2021)]{geva2021kv}
M.~Geva, R.~Schuster, J.~Berant, and O.~Levy.
\newblock Transformer feed-forward layers are key-value memories.
\newblock EMNLP 2021.

\bibitem[Li \& Liang(2021)]{li2021prefix}
X.~L. Li and P.~Liang.
\newblock Prefix-tuning: Optimizing continuous prompts for generation.
\newblock ACL 2021.

\bibitem[Zhang et~al.(2023)]{zhang2023adalora}
Q.~Zhang, M.~Chen, A.~Bukharin, et~al.
\newblock AdaLoRA: Adaptive budget allocation for parameter-efficient fine-tuning.
\newblock ICLR 2023.

\bibitem[Dettmers et~al.(2023)]{dettmers2023qlora}
T.~Dettmers, A.~Pagnoni, A.~Holtzman, and L.~Zettlemoyer.
\newblock QLoRA: Efficient finetuning of quantized LLMs.
\newblock NeurIPS 2023.

\end{thebibliography}

\appendix

\titleformat{\section}
  {\color{arxivink}\sffamily\large\bfseries}{Appendix~\thesection:}{0.6em}{}

\section{LOCOMO diagnostics: token-F1 vs.\ LLM-judge}
\label{app:locomo}

\begin{figure}[t]
  \centering
  \includegraphics[width=0.7\linewidth]{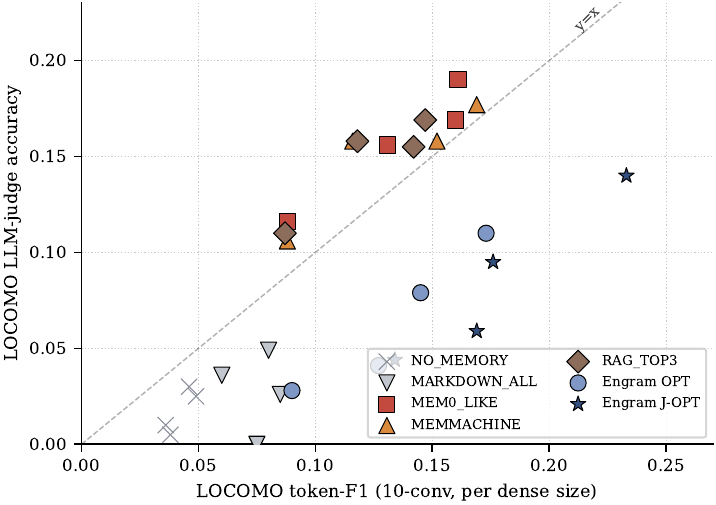}
  \caption{LOCOMO token-F1 vs.\ LLM-judge accuracy across all
  system $\times$ dense-size cells. Retrieval baselines
  (red/orange) sit near the $y{=}x$ line: their token-F1 and
  LLM-judge scores roughly agree. User-as-Engram OPT and Joint
  OPT (blue/cyan) sit systematically \emph{above} the line,
  indicating that token-F1 over-credits Engram. The asymmetry is
  the metric mismatch documented in Section~\ref{sec:locomo}.}
  \label{fig:metric-mismatch}
\end{figure}

\begin{figure}[t]
  \centering
  \includegraphics[width=0.82\linewidth]{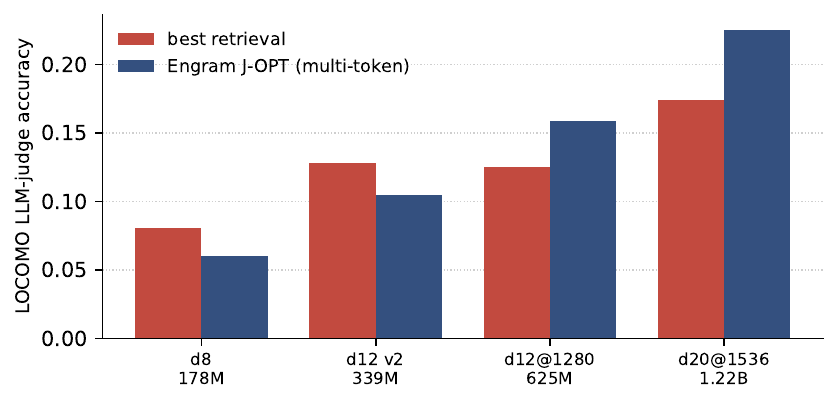}
  \caption{LOCOMO single-hop under an LLM judge (Qwen2.5-14B),
  matched multi-token pipeline. First-token OPT lets token-F1
  over-credit Engram for a correct first token over a noisy
  continuation (Figure~\ref{fig:metric-mismatch}); training the
  \emph{full} answer with multi-token Joint-OPT fixes this and
  overtakes the best retrieval baseline from d12@1280 upward
  ($+27$\% at 625\,M, $+29$\% at 1.22\,B). Below that cross-over
  the dense backbone is too small for the continuation to fall
  into place after the Engram-anchored prefix.}
  \label{fig:locomo-judge}
\end{figure}

The category breakdown (Figure~\ref{fig:locomo-categories}) and the
underlying per-cell token-F1 for every system, scale, and category
(Table~\ref{tab:locomo-categories}) are in the body
(Section~\ref{sec:locomo}). This appendix collects the two
diagnostic figures behind the LLM-judge discussion of
Section~\ref{sec:locomo}: Figure~\ref{fig:metric-mismatch}
shows that token-F1 over-credits Engram relative to an LLM judge,
and Figure~\ref{fig:locomo-judge} shows that multi-token Joint-OPT
closes the resulting judge gap.

\section{Mechanistic analysis}
\label{app:mech}

This appendix consolidates the mechanistic evidence that is
referenced piecewise in the main text (locality in
Section~\ref{sec:locality}, the multi-hop decomposition in
Section~\ref{sec:multihop}) into a single account of \emph{how} a
per-user fact becomes a retrievable, leak-free parametric edit.
The picture is deliberately simple: User-as-Engram is not an
opaque learned circuit but a \textbf{content-addressable memory}
whose every step is either deterministic (hashing, gating) or
directly measurable (the value the row writes, the gate it opens,
the locality of its effect). We trace the path of one inserted fact
through five observable stages, verifying each on the trained
Mini-Engrams (d12@1280 and d20, 16 canonical facts). All numbers
below are on the trained models unless stated.

\paragraph{(1) Addressing: a fact's trigger N-gram is a hash key.}
A fact is written at the row(s) that its trigger suffix N-gram
hashes to, via the deterministic multiplicative-XOR hash of the
Engram architecture~\citep{cheng2026engram}. The address is a
pure function of surface tokens---no learning, no per-user
state---which is why two users' edits never interfere unless
their \emph{distinguishing} tokens collide. A collision audit on
the randomly-initialized Engram demonstration
code~\citep{cheng2026engram} measures this directly: across
100 synthetic users $\times$ 30 facts, pairwise overlap on
user-distinguishing key tokens (\texttt{Patel}, \texttt{Portland},
\texttt{1991}) is just 4.5\%, and a 30-fact user occupies
0.027\% of the demo table (the table is 99.97\% empty after one
user). Addressing is sparse and content-keyed,
so capacity is never the limiting factor
(Section~\ref{sec:capacity-ablation}).

\paragraph{(2) The write opens its own gate and injects exactly its
value path.} A row reaches the residual stream only through the
gated value $\alpha_t\,W_V e$, so a write has to do two things, and we
can watch both. \emph{It opens its own gate.} Because the gate's key
is $W_K e$, writing the row raises the trigger-position gate from
$\alpha\!\approx\!0.015$ (an untrained synthetic trigger barely
fires) to $\alpha\!\approx\!0.99$ under the deployed OPT strategy
($0.59$ under closed-form UNEMBED\_P), while every non-trigger
position stays at $\alpha\!\approx\!0.03$--$0.04$ (identical on
d12@1280 and d20). \emph{It injects its value path.} The residual
change the deployed OPT row induces at the trigger has cosine
\textbf{0.998} (d12@1280) / \textbf{0.999} (d20) with the analytic
$W_V e$ projection---the gate and short-convolution scale the
injection but do not redirect it. This reproduces, on the trained
model and for the strategy we deploy, the read/write existence proof
on the random-init demonstration code (where writing a marker into
one trigger's rows moved the trigger position by $116.96$ vs.\ a mean
$0.65$ elsewhere, cosine $0.998$ to the predicted $W_V$ projection). (The small-magnitude
closed-form marker is dominated by the convolution nonlinearity, so
its \emph{direction} is noisier; OPT drives the row norm up, $99\to
132$, until the value path dominates.) The trained model corroborates
this from the behavioral side via the sensitivity asymmetry
(Section~\ref{sec:repro}): ablating the Engram pathway degrades \emph{factual} top-5 recall
(93.3\% retained) while \emph{reading} comprehension is untouched
(100\% retained)---a 6.7\,pp asymmetry that localises factual content
to the Engram pathway.

\paragraph{(2b) What the row \emph{encodes} vs.\ what it
\emph{optimizes}.} A subtler reading concerns the row's relation to
the gold token. The closed-form UNEMBED\_P row points its value path
straight at the gold token's unembedding (cosine \textbf{0.59}--%
\textbf{0.65} to $\text{lm\_head}[y]$): ``the row stores the gold
value'' is then literally true. Gradient OPT and Joint-OPT trade some
of that direct alignment (cosine drops to $0.16$--$0.24$) for higher
recall, shaping the \emph{full} next-token head rather than only the
gold unembedding direction. The interpretability is therefore in the
exact value-path injection (stage 2), not in the row equalling a
single unembedding vector---the deployed row is a value the network
reads cleanly, optimized against the true objective.

\paragraph{(3) Locality is exact---and it comes from addressing, not
from the gate.} It is tempting to attribute locality to the gate
(``it fires only on the trigger''), but the gate measurement above
shows the gate does \emph{not} single out an arbitrary synthetic
trigger before the write. Locality has a simpler source: we write
only the trigger's rows, and the trigger's suffix $N$-gram is unique
in the sequence, so every other position reads unchanged rows and its
output is bit-identical. We verify the consequence not on one fact but
across \emph{all 16 canonical facts and both insertion strategies}:
the maximum residual change over every
non-trigger position and every layer is \textbf{0.000}, and the change
before the Engram layer is \textbf{0.000}---exact, through every
subsequent attention/MLP layer (the trigger position itself moves by
order 1). Figure~\ref{fig:locality} contrasts this with a per-user
LoRA fit to the \emph{same} fact, whose effect is nonzero at every
position and every layer and which perturbs unrelated text by mean
$107$. Because the edit is \emph{addressed}, a per-user write
contaminates no unrelated forward pass ($\Delta$bpb $+0.00005$ vs.\ a
global LoRA's $+1.78$; Section~\ref{sec:negative}), and cross-user
leakage is zero by design (Section~\ref{sec:method}).

\begin{figure}[t]
  \centering
  \includegraphics[width=0.7\linewidth]{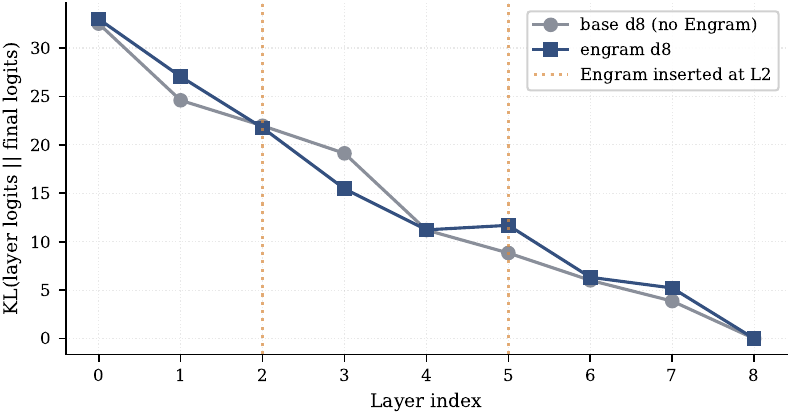}
  \caption{LogitLens KL by layer (Mini-Engram-d8 vs.\ base-d8).
  Engram converges faster at layer 3 (--3.66 KL gap), confirming
  \citeauthor{cheng2026engram}'s effective-deepening claim at our scale (stage 4
  of the mechanism above).}
  \label{fig:logitlens}
\end{figure}

\paragraph{(4) Depth: the edit lands where the model has already
``deepened''.} Engram rows are read at a late Engram layer, after
the network has effectively committed to a next-token
distribution. The LogitLens trace (Figure~\ref{fig:logitlens})
shows Mini-Engram-d8 converging to its output distribution
\emph{faster} than the base---layer-3 KL is 3.66 lower---%
reproducing \citeauthor{cheng2026engram}'s ``effective deepening'' at our scale.
A row therefore overrides a near-final prediction at exactly the
position where it is decoded, rather than steering an early,
still-malleable representation (the place where a global LoRA
change does its damage). We make this dependence \emph{causal}
rather than merely correlational by re-running the insertion at the
model's \emph{early} Engram layer instead of the late one (layer 2
vs.\ layer 11 on d20, layer 2 vs.\ 7 on d12@1280; identical OPT-15
budget): top-1 recall collapses from
\textbf{1.00 at the late layer to 0.25} (d20) / \textbf{0.31}
(d12@1280) \textbf{at the early layer}, because an early injection
must survive the rest of the transformer stack and largely washes
out. The depth at which the Engram reads is not incidental---it is
where a value-path edit can override the prediction.

\paragraph{(5) Limit: the gate is a surface-N-gram hash, so it
cannot compose across triggers.} The same mechanism that makes
the edit local also bounds what it can do. The multi-hop
decomposition (Section~\ref{sec:multihop},
Figure~\ref{fig:multihop}) is the cleanest mechanistic readout in
the paper: on a balanced 63-pair corpus, chained queries succeed
\textbf{91\%} of the time when the query's suffix N-gram
\emph{overlaps} the second fact's trigger, but only \textbf{13\%}
(${\lesssim}10\%$ after filtering first-fact-wins coincidences)
when true composition is required---even though per-fact direct
recall is 99.2\% in both subsets. The near-bimodal split is exactly
what a surface-keyed lookup predicts: the gate fires on a matching
trigger and returns the stored token, or it does not fire and the
gold token is far from the top. There is no chaining circuit to
find because, by design, there is none.

\paragraph{Synthesis.} The five stages compose into one sentence:
\emph{User-as-Engram hashes a fact's surface trigger to a sparse
row whose write both opens its own gate ($\alpha\,0.02\to0.99$) and
injects exactly its value path (cosine 0.999 to the predicted
projection), at a late, already-deepened layer (early insertion
recall collapses 1.00$\to$0.25), with the change exactly 0.000 at
every other position---and therefore retrieves single facts with
zero contamination but cannot compose across facts.} This is why the
method cleanly separates \emph{content} (the rows) from
\emph{reasoning skill} (composition, which must come from the shared
LoRA in the layered design of Section~\ref{sec:layered})---the
content/skill split the paper proposes is not an engineering
convenience but a direct consequence of the mechanism.

\section{Pretraining curves}
\label{app:pretrain}

\begin{figure}[t]
  \centering
  \includegraphics[width=\linewidth]{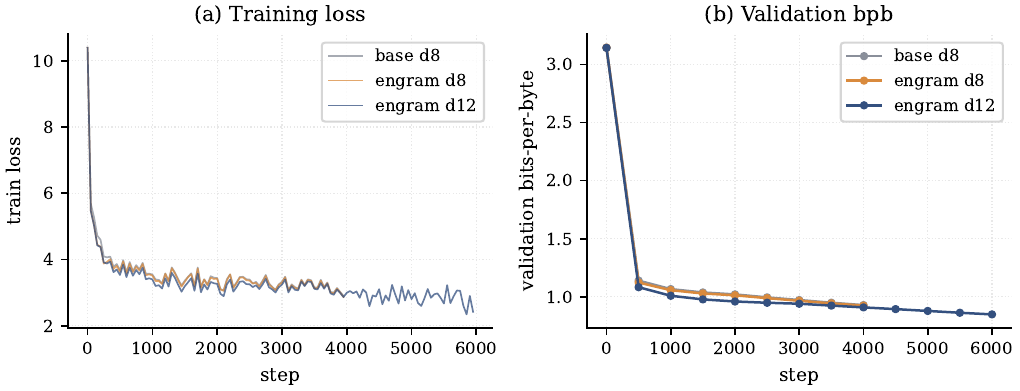}
  \caption{Mini-Engram pretraining curves. Engram d8 reaches lower
  validation bpb than base d8 at iso-FLOPs (matching the paper's
  finding at much smaller scale); engram d12 is substantially better
  thanks to the larger backbone and longer training.}
  \label{fig:pretrain}
\end{figure}

Figure~\ref{fig:pretrain} shows the Mini-Engram pretraining runs that
every later insertion experiment builds on. Adding the Engram table is
not a tax on language modeling: at iso-FLOPs the d8 Engram model
reaches marginally lower validation bits-per-byte than the d8 baseline
(0.924 vs.\ 0.929 at step 4\,000), reproducing
\citeauthor{cheng2026engram}'s effective-deepening finding at our much
smaller scale, and the larger d12 Engram model---more backbone, longer
training---is substantially better (0.849 bpb). These checkpoints are
the base models for the row-insertion results in the body.

\section{Insertion strategy comparison}
\label{app:strategies}

\begin{figure}[t]
  \centering
  \includegraphics[width=0.7\linewidth]{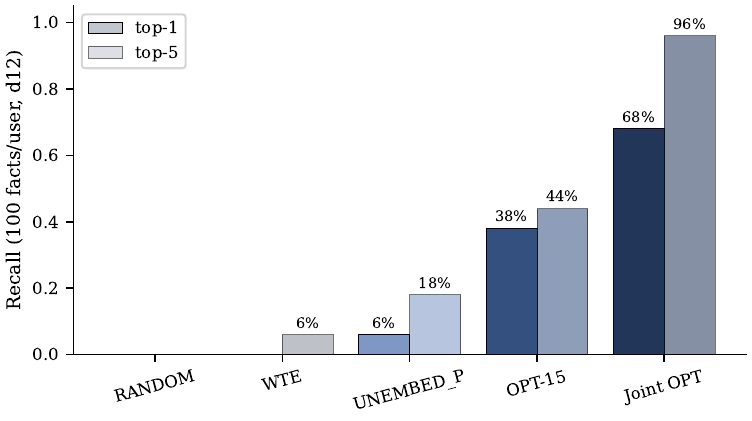}
  \caption{Top-1 and top-5 recall at 100 facts/user, Mini-Engram-d12.
  RANDOM (control) confirms our signal isn't noise; UNEMBED\_P alone
  is too weak; OPT-15 helps; Joint OPT closes most of the gap to
  full LoRA fine-tuning at 161$\times$ less storage.}
  \label{fig:strategies}
\end{figure}

Figure~\ref{fig:strategies} compares five ways of writing a fact
into Engram rows at the hardest density we test (100 facts/user,
Mini-Engram-d12). A random marker (RANDOM, a control) and the token
input embedding (WTE) never recover the fact (0\% top-1), confirming
the recall signal is not an artefact of the metric. The closed-form
pseudo-inverse marker (UNEMBED\_P) reaches only 6\%/18\% top-1/top-5:
it orients the row correctly but the gate suppresses its magnitude.
Fifteen steps of per-row gradient descent (OPT-15) lift this to
38\%/44\%, and Joint OPT---optimizing all touched rows together
against the user's full fact set---reaches 68\%/96\%, recovering most
of what a per-user LoRA achieves at this density
(Figure~\ref{fig:density}) at $161\times$ less storage. The earlier
single-fact-at-a-time comparison on the original 16-fact USER+ORG
benchmark (Mini-Engram-d8) is reported in
Appendix~\ref{app:strategies-full}.

\section{Paraphrase generalization}
\label{app:paraphrase}

\begin{figure}[t]
  \centering
  \includegraphics[width=0.75\linewidth]{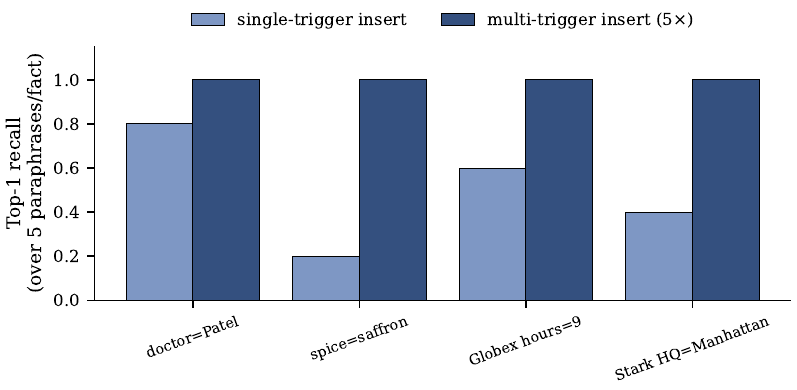}
  \caption{Paraphrase generalization across 4 facts $\times$ 5 paraphrases
  each. Single-trigger insertion gets 50\% top-1 for free (suffix
  N-gram overlap); multi-trigger insertion (insert at all 5
  paraphrases) gets 100\% at 5$\times$ the per-fact OPT cost.}
  \label{fig:paraphrase}
\end{figure}

Figure~\ref{fig:paraphrase} tests whether a fact written at one
trigger phrasing answers the same question asked differently, across
five held-out paraphrases per fact. Single-trigger insertion (write
once, at the canonical phrasing) averages 50\% top-1: 4/5 for the
``doctor'' fact, 3/5 for ``Globex hours'', 2/5 for ``Stark HQ'', and
1/5 for ``spice''. Generalization tracks suffix $N$-gram overlap---
paraphrases that end in the same tokens as the trained trigger
transfer for free, while divergent surface forms (e.g.\ ``I love the
spice'' vs.\ ``my favorite spice is'') do not. Multi-trigger
insertion (write the fact at all five phrasings, at $5\times$ the
per-fact OPT cost) drives every paraphrase to top-1. The full
per-paraphrase rank data are tabulated in
Appendix~\ref{app:paraphrase-detail}.

\section{Multi-domain composition}
\label{app:multidomain}

\begin{figure}[t]
  \centering
  \includegraphics[width=0.8\linewidth]{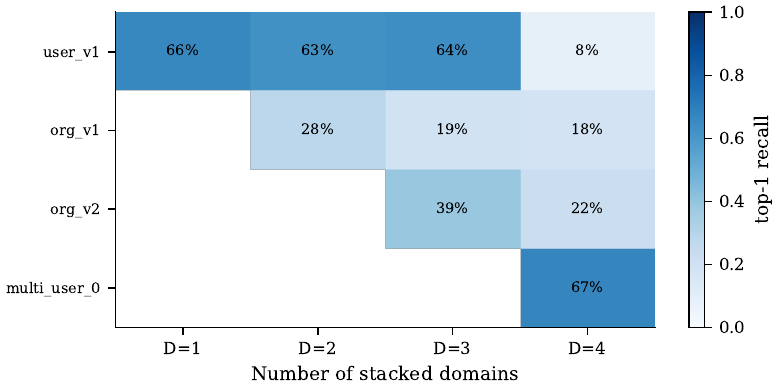}
  \caption{Multi-domain additive composition heatmap. Each cell shows
  per-domain top-1 recall when $D$ domains are stacked. Heavy
  degradation when domains share trigger templates (high address
  overlap). Disjoint-template domains (corp + user, demonstrated in
  main text) compose without loss.}
  \label{fig:multidomain}
\end{figure}

Figure~\ref{fig:multidomain} stacks $D$ fact domains into one Engram
table and reports per-domain top-1 as $D$ grows. The user domain
(\texttt{user\_v1}) is unaffected by adding disjoint-template
corporate domains: it holds 66\%\,$\to$\,63\%\,$\to$\,64\% as
\texttt{org\_v1} and \texttt{org\_v2} are layered in ($D{=}1$ to 3).
The collapse appears at $D{=}4$, when a second \emph{user-template}
domain (\texttt{multi\_user\_0}) is added: because it reuses the same
trigger templates, the two user domains collide in the address hash
and \texttt{user\_v1} falls to 8\% while the newcomer reads 67\%.
Domains that share templates interfere (the corporate domains also
degrade each other, \texttt{org\_v1} 28\%\,$\to$\,18\%), whereas
template-disjoint domains compose without loss---consistent with the
corp+user composition measured directly in Section~\ref{sec:additive}.

\section{Serving latency CDF}
\label{app:cdf}

\begin{figure}[t]
  \centering
  \includegraphics[width=0.75\linewidth]{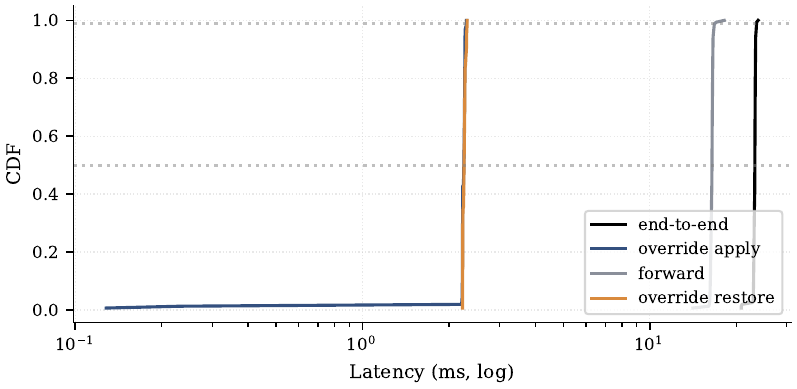}
  \caption{Per-request latency CDF on Mini-Engram-d12 (30 users
  $\times$ 50 facts $\times$ 600 requests). Override apply (blue) and
  restore (orange) are both sub-3 ms at p99; the forward pass (grey)
  dominates total latency (black).}
  \label{fig:cdf}
\end{figure}

\begin{figure}[t]
  \centering
  \includegraphics[width=\linewidth]{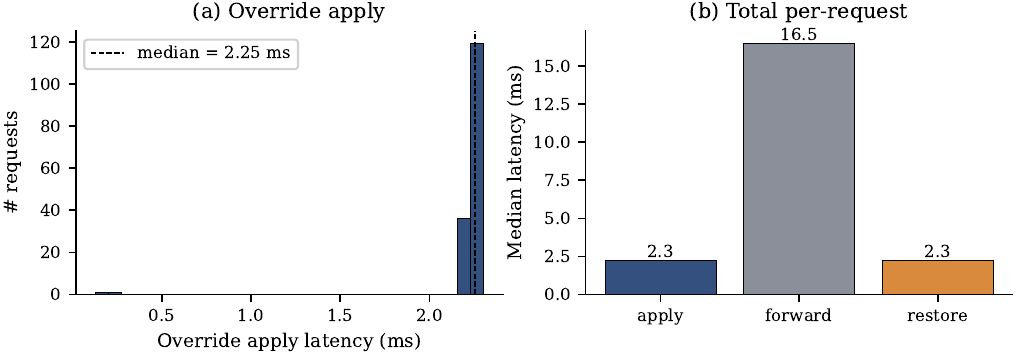}
  \caption{Per-request latency breakdown, Mini-Engram-d12 (30 users
  $\times$ 50 facts $\times$ 600 requests). \textbf{(a)} Override-apply
  latency: median 2.2 ms. \textbf{(b)} Median per-request latency by
  component: apply (2.3 ms) + forward (16.5 ms) + restore (2.3 ms);
  the frozen forward pass dominates, and the override apply+restore
  overhead ($\sim$4.6 ms) is under 20\% of the 23.2 ms end-to-end
  median.}
  \label{fig:serving}
\end{figure}

Figure~\ref{fig:cdf} shows the per-request latency distribution under
multi-tenant serving (the earlier \texttt{d12}~v2 checkpoint, 339\,M,
30 users $\times$ 50 facts, 600 single-token recall probes), and
Figure~\ref{fig:serving} breaks the median request into its three
components. Applying a user's override map and restoring the base
table afterwards are both cheap and tightly bounded---median
2.2\,ms, p99 2.3\,ms each---so the swap adds no latency tail. The
forward pass dominates: a median of 16.5\,ms out of a 23.2\,ms
end-to-end total (p99 27.8\,ms), so per-user memory adds under 20\%
overhead on top of a request the model would run anyway. These
absolute numbers are higher than the ablation-optimal
\texttt{d12@1280} serving run reported in
Figure~\ref{fig:serving-scale} (0.03\,ms apply, 4.4\,ms p50,
226\,req/s on an idle GPU)---both the apply cost and the forward pass
are faster on that newer benchmark---but the qualitative conclusion is
the same in both: the apply/restore swap is a small fraction of
per-request latency, and per-request work is independent of how many
tenants share the server.

\section{Qwen-3B + RAG details}
\label{app:qwen-rag-prompt}

The Qwen-3B + RAG measurements in Section~\ref{sec:qwen-rag} use
\texttt{Qwen/Qwen2.5-3B-Instruct} in bf16 with greedy decoding
($\le 32$ new tokens). The chat-template prompt is:

{\small
\begin{verbatim}
[SYSTEM]
You are answering personal-memory questions about the user. Use only
the facts provided. Answer with the shortest possible span (typically
1-3 words). Do not explain or add any extra text.

[USER]
Facts about the user:
- <retrieved fact 1>.
- <retrieved fact 2>.
...

Question: <indirect question from indirect_qa>
Answer:
\end{verbatim}}

Retrieval uses the same \texttt{all-MiniLM-L6-v2}
encoder as the RAG/\allowbreak MEM0/\allowbreak MEMMACHINE baselines of
Section~\ref{sec:memory-systems}, with the per-user fact set as the
index. For each indirect probe the question text is embedded and
cosine-similarity top-$k$ retrieval returns the facts to put in the
context block. The same 20-user/20-probe split as
Figure~\ref{fig:layered-conditions} is evaluated.

\begin{figure}[t]
  \centering
  \includegraphics[width=0.9\linewidth]{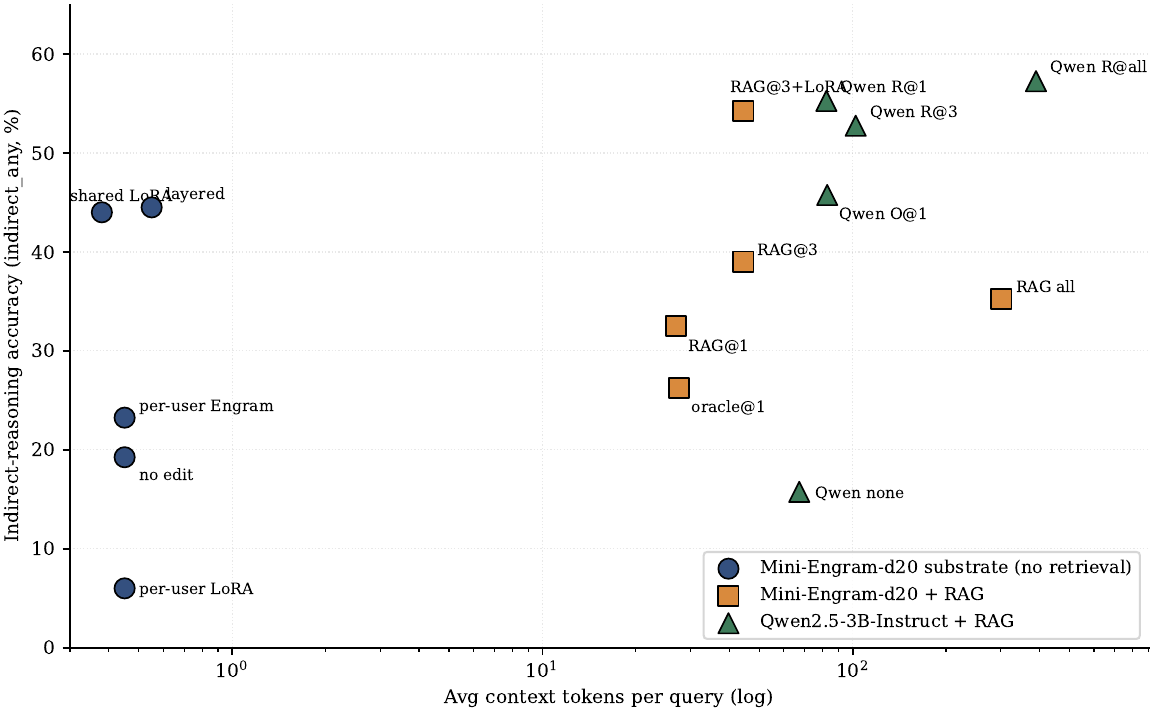}
  \caption{Indirect-reasoning accuracy (\texttt{indirect\_any}) vs.\ avg
  context tokens per query on the 20-user split. The layered design
  and the shared LoRA alone anchor the 0-context end; RAG
  top-3 + shared LoRA lifts accuracy to 54\% at 44 context tokens;
  Qwen-3B + RAG reaches 55--57\% at 82--390 tokens on
  a $\sim$2.5$\times$ larger backbone. Naive RAG on the same Mini-Engram-d20
  base sits \emph{below} the layered design: putting facts in
  context is not the same as having the reasoning skill.}
  \label{fig:pareto-rag}
\end{figure}

Two answer-match criteria are reported: \emph{indirect\_top1} requires
the first generated word to equal (case-insensitive) the first word of
the gold answer; \emph{indirect\_any} requires the gold span to occur
anywhere in the generation. This matches the substring matcher used
for the layered design.

\paragraph{Why oracle-top-1 underperforms RAG top-1.}
A non-obvious result in the Qwen-3B + RAG experiment is that
oracle top-1 retrieval (using the known required facts to
fetch the ground-truth fact) gives 46\% indirect\_any,
\emph{below} RAG top-1's 55\%. Inspecting the data, retrieval
often returns facts that are \emph{adjacent} to the required
key in semantic space and that happen to also contain the
answer surface (e.g.\ retrieving the ``birth-year'' fact when
asked about ``current age'' covers both the year and the
implicit age). RAG with $k=1$ thus over-samples the information
needed for indirect reasoning, while strict oracle retrieval
leaves the LM to derive the answer from one single-attribute
fact---which Qwen-3B often refuses to do under our ``shortest
span'' system prompt.

\section{Alternative architectures we considered}
\label{app:alt}

For completeness we report two alternative designs we benchmarked
before settling on per-user override tables.

\paragraph{Shared table with per-user hash salt (deprecated).}
In this design, every user's facts coexist in one physical table;
their hashes are XOR-salted by user-id so identical surface
triggers map to disjoint rows. We benchmark on Mini-Engram-d12
(Table~\ref{tab:salt-design}):

\begin{table}[t]
  \centering
  \caption{Shared-table-with-salt: own-fact recall vs.\ leak rate.
  All numbers on Mini-Engram-d12 with the XL corpus.}
  \label{tab:salt-design}
  \begin{tabular}{lcc}
    \toprule
    config & own-fact top-1 & cross-user leak \\
    \midrule
    no salt $+$ UNEMBED\_P (cheap, 100$\times$100 facts) & 0.089 & 0.039 \\
    salt $+$ OPT-15 (10$\times$30 facts) & 0.11 & \textbf{0.000} \\
    salt $+$ strong OPT-60 (10$\times$30) & 0.46 & \textbf{0.000} \\
    salt $+$ strong OPT-60 (30$\times$100) & 0.345 & 0.005 \\
    \bottomrule
  \end{tabular}
\end{table}

The salt design provides architectural privacy at the cost of
recall: salted addresses point to untrained rows (the model
never saw salt during pretraining), so the gate fires more
weakly and OPT must work harder. Because per-user override
tables (main body Section~\ref{sec:serving}) provide stronger
privacy by design \emph{and} use the trained address
space, we recommend overrides as the production design and keep
salt as a fallback for cases where the storage backend cannot
afford per-user override maps.

\paragraph{100-user serving: UNEMBED\_P vs.\ Joint OPT.}
At 100 users $\times$ 100 facts on Mini-Engram-d12@1280, the
cheap (closed-form) UNEMBED\_P strategy reaches 9.2\% top-1 /
23.4\% top-5 at 2.3\,ms apply latency. Joint OPT (1000 steps,
the production setting) on the same configuration reaches
\textbf{62\% top-1 / 96\% top-5} at \textbf{226\,req/s and
4.4\,ms p50 latency} on an idle GPU
(Figure~\ref{fig:serving-scale}), matching the $\sim$60\%/95\%
prediction extrapolated from the 30-user data and consistent
with the per-user-density curve at d12@1280
(Section~\ref{sec:density}: 67\% top-1 / 99\% top-5
at 100 facts/user). The serving-protocol cross-user leak under
Joint OPT is 6.1\% from gold-value coincidence (7 of 114
cross-user probes), versus 46\% under the all-coresident
protocol where every user's overrides occupy the same table at
once.

\section{Insertion strategies (full)}
\label{app:strategies-full}

Table~\ref{tab:strategies-full} reports the first comparison of
insertion strategies on Mini-Engram-d8 with
the original 16-fact USER+ORG benchmark (single-fact-at-a-time):

\begin{table}[t]
  \centering
  \caption{Insertion strategy comparison on the original 8 USER + 8 ORG
  benchmark, Mini-Engram-d8. RANDOM is a control; UNEMBED\_P is the
  closed-form pseudo-inverse; OPT is 15-step gradient on the row.
  Numbers are mean across the 16 facts.}
  \label{tab:strategies-full}
  \begin{tabular}{lccccc}
    \toprule
    strategy & mean $\Delta$logit & $+\Delta$logit & rank improved & top-1 hits & top-5 hits \\
    \midrule
    RANDOM   & --4.708 & 1/16 & 2/16 & 0/16 & 0/16 \\
    WTE      & --0.150 & 6/16 & 6/16 & 0/16 & 1/16 \\
    UNEMBED\_P & +1.711 & 15/16 & 13/16 & 1/16 & 3/16 \\
    \textbf{OPT (15 steps)} & \textbf{+5.323} & \textbf{16/16} & \textbf{15/16} & \textbf{6/16} & \textbf{7/16} \\
    \bottomrule
  \end{tabular}
\end{table}

The DUAL strategy (jointly satisfying gate-K and value-V via
least squares) was tested, but proved \emph{worse} than
UNEMBED\_P (4/16 $+\Delta$logit)---over-constraining $e$ drives
it off the trained-row distribution and the gate suppresses it.
We dropped DUAL.

\section{Per-paraphrase rank table}
\label{app:paraphrase-detail}

Table~\ref{tab:paraphrase} gives the full per-paraphrase rank data
behind Section~\ref{sec:paraphrase}
and Figure~\ref{fig:paraphrase}, single-trigger insertion on
Mini-Engram-d8:

\begin{table}[t]
  \centering
  \caption{Per-paraphrase rank of the gold token after single-trigger
  OPT insertion. ``$\star$'' marks rank 0 (top-1).}
  \label{tab:paraphrase}
  \footnotesize
  \begin{tabular}{lll}
    \toprule
    insertion trigger & query (paraphrase) & rank \\
    \midrule
    My doctor's name is Dr. & My doctor's name is Dr. & 0 $\star$ \\
    & My doctor is Dr. & 0 $\star$ \\
    & Who is my doctor? Dr. & 54 \\
    & My physician's name is Dr. & 0 $\star$ \\
    & My GP is Dr. & 0 $\star$ \\
    \midrule
    My favorite spice is & My favorite spice is & 0 $\star$ \\
    & I love the spice & 1399 \\
    & My preferred spice is & 32 \\
    & The spice I like most is & 45 \\
    & My go-to spice is & 94 \\
    \midrule
    Globex office hours start at & Globex office hours start at & 0 $\star$ \\
    & Globex opens at & 1 \\
    & Globex starts work at & 0 $\star$ \\
    & What time does Globex open? At & 2 \\
    & Globex's day begins at & 0 $\star$ \\
    \midrule
    Stark Industries headquarters is in & Stark Industries headquarters is in & 0 $\star$ \\
    & Stark HQ is in & 1949 \\
    & Stark is based in & 216 \\
    & Stark Industries is located in & 66 \\
    & The Stark headquarters is in & 0 $\star$ \\
    \bottomrule
  \end{tabular}
\end{table}

Generalization is high when paraphrases share suffix N-grams
(e.g., all ``doctor'' paraphrases end in ``Dr.''); poor when
the surface form diverges (``I love the spice'' vs.\ ``My
favorite spice is''). Inserting at all 5 paraphrases per fact
(multi-trigger) drives every paraphrase to rank 0.

\section{Joint OPT convergence}
\label{app:joint-loss}

\begin{figure}[t]
  \centering
  \includegraphics[width=0.7\linewidth]{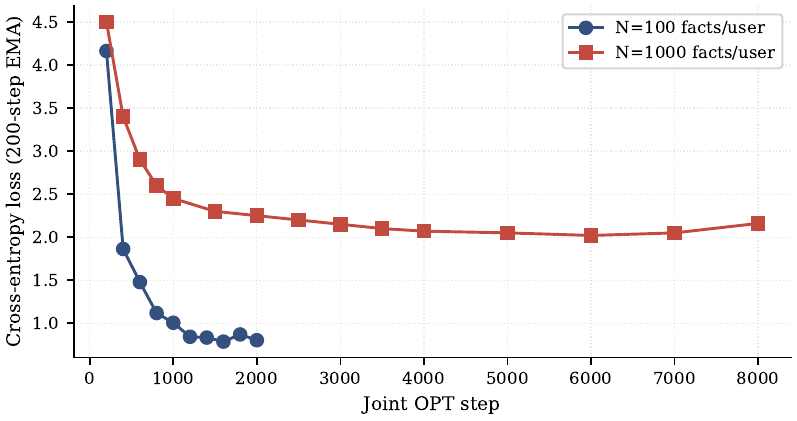}
  \caption{Joint OPT convergence on Mini-Engram-d12. At 100 facts/user,
  loss converges to 0.8 within 1500 steps; at 1000 facts, loss
  saturates around 2.0 due to higher within-user fact-density
  interference (Section~\ref{sec:density}).}
  \label{fig:jointopt}
\end{figure}

Figure~\ref{fig:jointopt} traces the Joint OPT optimization behind the
density curves of Section~\ref{sec:density}. At 100 facts/user the joint
objective converges to $\approx$0.8 cross-entropy within $\sim$1\,500
steps; at 1\,000 facts/user it plateaus near 2.0, the higher floor
reflecting the within-user address interference that caps recall at high
density. Convergence is monotone and stable in both regimes, so the
fixed step budgets we use (2\,000 steps at 100 facts, 8\,000 at 1\,000)
leave no easy gains on the table.

\section{Joint OPT algorithm}
\label{app:joint-opt}

\begin{enumerate}[leftmargin=*,topsep=2pt,itemsep=1pt]
  \item For each fact $f$: compute trigger global rows $R_f$ (deterministic from tokens) and UNEMBED\_P initial marker $e_f^{(0)} = W_V^\dagger U_{y_f}$.
  \item Allocate one trainable tensor over the union of all touched
        rows, $\mathbf{R} = \cup_f R_f$. Initialize: each row
        $\mathbf{R}[i] = \mathrm{mean}_{f: i \in R_f} e_f^{(0)}[i]$
        (average of UNEMBED\_P rows where multiple facts share an
        address).
  \item Zero out the embedding rows in $\mathbf{R}$. Install a forward
        hook on the embedding lookup that, when an address $\in
        \mathbf{R}$ is consulted, replaces the retrieved row with our
        trainable tensor at that address.
  \item For $K$ steps: sample a random fact $f$, compute the
        cross-entropy loss on $y_f$ at the trigger position,
        backprop only to $\mathbf{R}$.
  \item After training, write $\mathbf{R}$ into the embedding table
        as the user's override map.
\end{enumerate}

We use $K = 2000$ for 100 facts and $K = 8000$ for 1000 facts, Adam
LR 0.5. Wall time scales as $K$ alone (one fwd+bwd per step) so per-fact
cost \emph{decreases} with more facts: 0.44 s/fact at $N{=}100$,
0.23 s/fact at $N{=}1000$.

\section{Seed variance of the layered measurement}
\label{app:layered-seeds}

The headline layered-design vs.\ per-user-LoRA numbers in
Figure~\ref{fig:layered-conditions} are
from a single LoRA training run; for transparency, we re-ran the same
configuration with two further RNG states and compared
(Table~\ref{tab:layered-seeds}). Because the
training-data ordering, LoRA-weight initialization, and Adam-state
warmup are stochastic, per-user val\_bpb $\Delta$ varies across seeds
even with identical hyperparameters.

\begin{table}[h]
  \centering
  \caption{Three-seed variance on the layered-design vs.\
  per-user-LoRA contrast,
  Mini-Engram-d20, $n{=}20$ users per seed, identical config across
  seeds (rank-64 LoRA, 1500 steps, lr 5e-4, lora\_alpha 128;
  per-user Engram J-OPT 1500 steps lr 0.5; eval\_tokens 524288;
  only seeds differ in LoRA-weight initialization and Adam-state /
  data-ordering RNG). The architectural property (per-user Engram
  $\Delta$bpb at noise floor $\sim$$10^{-5}$) is seed-invariant; the
  LoRA contamination magnitude varies by $\sim$13\% and the layered
  design's indirect\_any varies by $\sim$10\,pp across seeds while the
  qualitative ranking (layered $\gg$ LoRA on indirect, LoRA $\gg$
  layered on contamination) is preserved in every seed.}
  \label{tab:layered-seeds}
  \resizebox{\linewidth}{!}{%
  \begin{tabular}{lcccc}
    \toprule
    quantity & seed S0 & seed S1 & seed S2 & 3-seed mean (range) \\
    \midrule
    per-user LoRA $\Delta$bpb mean       & $+1.784$ & $+1.754$ & $+1.557$ & $+1.698$\ {[}+1.56, +1.78{]} \\
    per-user LoRA $\Delta$bpb per-user range & {[}+0.44,\,+3.74{]} & {[}+0.73,\,+2.46{]} & {[}+0.69,\,+2.89{]} & --- \\
    per-user Engram $\Delta$bpb mean       & $+0.000052$ & $+0.000049$ & $+0.000054$ & $+0.000052$ \\
    LoRA+Engram stack $\Delta$bpb mean       & $+1.819$ & $+1.756$ & $+1.290$ & $+1.622$\ {[}+1.29, +1.82{]} \\
    layered design $\Delta$bpb mean       & $+0.386$ & $+0.388$ & $+0.379$ & $+0.385$\ {[}+0.38, +0.39{]} \\
    layered design indirect\_any          & 44.5\% & 44.2\% & 34.8\% & 41.2\%\ {[}34.8, 44.5{]}\,\% \\
    per-user LoRA indirect\_any          & 6.0\% & 8.8\% & 7.2\% & 7.3\%\ {[}6.0, 8.8{]}\,\% \\
    layered design direct top-1           & 100\% & 100\% & 100\% & 100\% \\
    per-user LoRA direct top-1           & 99\% & 97.5\% & 100\% & 98.8\%\ {[}97.5, 100{]}\,\% \\
    layered design worse than base           & 0/20 & 0/20 & 0/20 & \textbf{0/60} \\
    per-user LoRA worse than base           & 17/20 & 16/20 & 16/20 & \textbf{49/60 (82\%)} \\
    LoRA+Engram stack worse than base           & 18/20 & 16/20 & 16/20 & \textbf{50/60 (83\%)} \\
    \bottomrule
  \end{tabular}}
\end{table}

Reading: the architectural-property comparison (the per-user
Engram's $\Delta$bpb $\ll$ per-user LoRA's by $\sim$33{,}000$\times$) is
seed-stable to four decimal places---every seed's Engram is at the
noise floor and every seed's LoRA is in $[+1.56, +1.78]$. The
conclusion that the layered design beats per-user LoRA on every
measure is also seed-stable: the layered design
is never worse than the no-edit base on indirect (0/60 across seeds),
while per-user LoRA is worse in 49/60 = 82\%. The exact magnitudes vary
modestly: seed-to-seed layered indirect\_any spans $34.8$\%--$44.5$\%
(mean $41.2$\%), and the layered-vs-LoRA indirect ratio spans $4.8\times$
(S2) to $7.4\times$ (S0). The body headlines the canonical seed
S0 ($7.4\times$, 44\% vs.\ 6\%) and cites the 3-seed mean
($5.6\times$, 41\% vs.\ 7\%) alongside it; this appendix makes the
per-seed variance explicit so readers can see which numbers depend
on which seed.

\section{Extended experiments and ablations}
\label{app:extended}

This appendix collects the capacity, fact-count, dense-size, and
rank ablations, and the layered-architecture robustness studies,
that the main text summarizes in figures.

\subsection{Multi-fact-in-the-loss finetune, and the teacher-trace negative result}
\label{sec:mf-finetune}
\begin{figure}[t]
  \centering
  \includegraphics[width=0.56\linewidth]{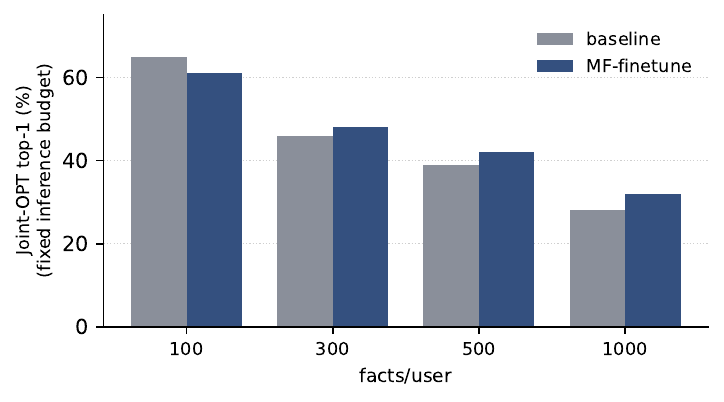}
  \caption{Multi-fact-in-the-loss finetune vs.\ baseline d12@1280
  on the Joint-OPT density curve (matched OPT-step schedule). At
  a fixed inference budget MF improves recall at high density
  ($+14$\% relative at $n{=}1000$, 8k steps). With the budget
  unpinned (12k steps) both reach $\sim$34.5\% at $n{=}1000$: MF
  \emph{accelerates convergence} rather than raising the
  asymptotic ceiling. val\_bpb is preserved (0.774 vs.\ 0.770).}
  \label{fig:mf-finetune}
\end{figure}

We test the multi-fact-in-the-loss recipe via a continuation
finetune of d12@1280 that injects many synthetic foreign rows per
batch and adapts the gate, value projection, and dense backbone to
handle them, while general LM ability is preserved (val\_bpb
unchanged). At a fixed inference budget it improves recall at high
density ($+14$\% relative at $n{=}1000$, 8k steps); with the budget
unpinned (12k steps) both the baseline and MF reach $\sim$34.5\% at
$n{=}1000$ (Figure~\ref{fig:mf-finetune}). So MF accelerates Joint-OPT
convergence rather than raising the asymptotic ceiling---an economic
win, not an architectural one. The teacher-trace variants
(245 Qwen3-8B~\citep{qwen2025qwen3} traces of
observation/statement/QA/$\langle$think$\rangle$-reasoning) cut
contamination ($\Delta$bpb $+0.386\to+0.233$/$+0.196$) but lowered
indirect\_any ($44.5\%\to38.5\%$/$28.0\%$; bootstrap 95\% CIs
strictly negative), because that trace format scores a
$\langle$think$\rangle$ block while our eval reads only the top
answer token.

\subsection{Engram capacity ablation}
\label{sec:capacity-ablation}

\begin{figure}[t]
  \centering
  \includegraphics[width=\linewidth]{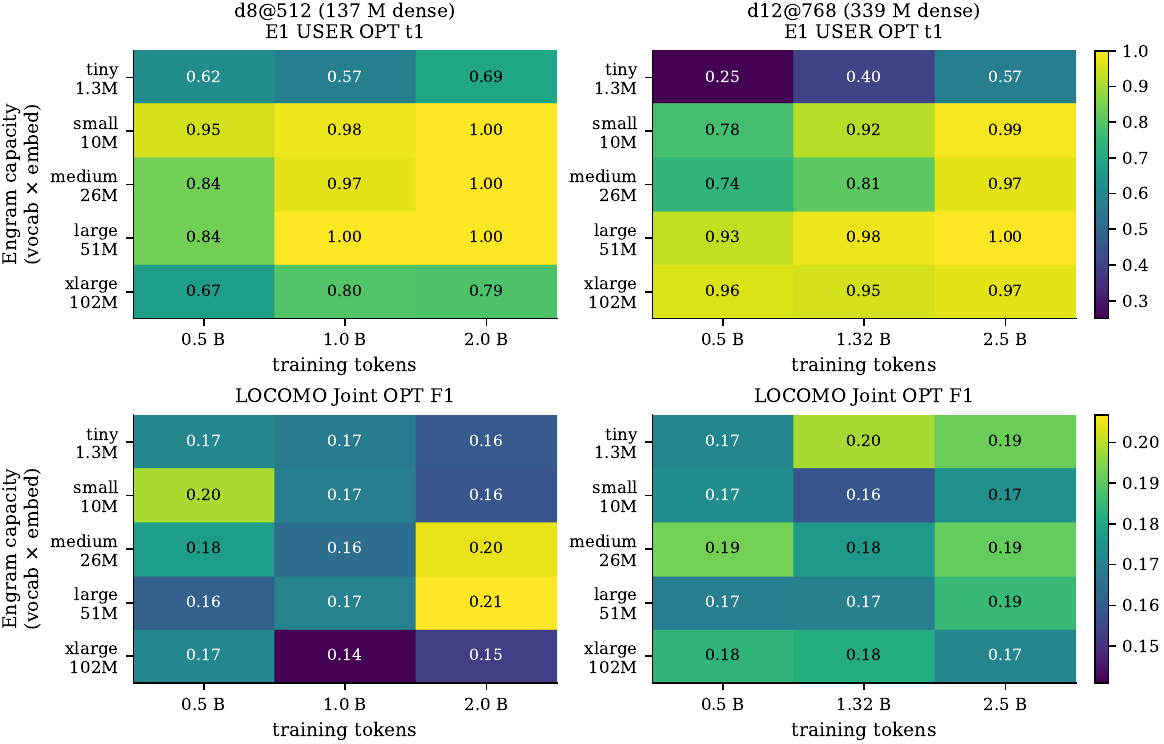}
  \caption{Engram capacity $\times$ token-budget response surface, at
  Mini-Engram-d8 v2 (left) and d12 v2 (right). Top row:
  E1 USER OPT top-1 (100-fact recall). Bottom row: LOCOMO Joint
  OPT F1. The same \texttt{large} (50\,K $\times$ 256) Engram
  size wins at the highest token budget for \emph{both} dense
  scales; \emph{tiny} is under-capacity, \emph{xlarge} is
  over-provisioned. The optimum scales in tokens, not capacity.}
  \label{fig:capacity-heatmap}
\end{figure}

How does Engram-table size interact with token budget at a fixed
dense backbone? We probe this as a $5 \times 3 \times 2$
matrix: five capacity levels ($v$ slots per
N-gram, $e$ total embed-dim per N-gram; Table~\ref{tab:capacity-levels}),
three token budgets, two
dense sizes (Figure~\ref{fig:capacity-heatmap}).

\begin{table}[t]
  \centering
  \caption{Capacity levels used in the ablation. Total Engram
  params $= 4 \cdot v \cdot e$ (2 N-gram orders $\times$ 2 layers).}
  \label{tab:capacity-levels}
  \begin{tabular}{lccc}
    \toprule
    label & $v$ slots & $e$ embed & total Engram \\
    \midrule
    tiny    & 5\,K   & 64  & 1.28\,M \\
    small   & 20\,K  & 128 & 10.24\,M \\
    medium  & 50\,K  & 128 & 25.6\,M \\
    large   & 50\,K  & 256 & 51.2\,M \\
    xlarge  & 100\,K & 256 & 102.4\,M \\
    \bottomrule
  \end{tabular}
\end{table}



\paragraph{Three regularities.}
(i) \emph{Capacity has an optimum.} Both endpoints
under-perform: \emph{tiny} cannot address enough distinct facts;
\emph{xlarge} under-trains each slot.
(ii) \emph{The optimum is the same Engram size at both dense
scales}: \texttt{large} (50\,K $\times$ 256 $=$ 51.2\,M params)
wins at the highest token budget for both d8 and d12.
\textbf{The optimum scales in tokens, not in capacity.}
(iii) \emph{Below the catch-up token budget, smaller Engrams
win.} At d8 with 0.5\,B tokens, \emph{small} (10\,M) beats
\emph{large} (51.2\,M) on ins-OPT 1.00 vs.\ 0.62---the larger
table is under-trained per slot. The cross-over occurs around
1\,B tokens for d8 and 1.32\,B tokens for d12.

\paragraph{Practical rule.} For an Engram pretrained at Karpathy
12 t/p of scaling-params, use the \texttt{large} (50\,K $\times$
256) Engram table from d12@768 onward. For d8-class models with
${\leq}\,1$\,B training tokens, use \texttt{small} (20\,K
$\times$ 128) instead. Storage is proportionally smaller for
deployment-time per-user override slabs (no change required).

\subsection{Fact-count scaling: per-fact independent OPT to 1000 facts}
\label{sec:factscale}

\begin{figure}[t]
  \centering
  \includegraphics[width=\linewidth]{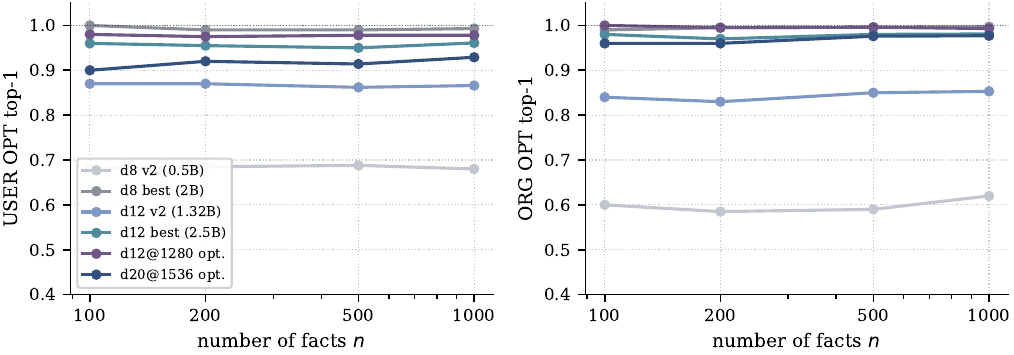}
  \caption{Per-fact independent OPT recall vs.\ fact count $n$.
  Left: USER OPT top-1. Right: ORG OPT top-1. The d12@1280
  and d20@1536 optimal cells (green, purple) hold near-1.00
  recall from $n=100$ to $n=1000$. No ceiling visible in this
  per-fact independent-OPT setting.}
  \label{fig:factscale}
\end{figure}

Once Engram capacity and tokens are chosen at the ablation
optimum, how does User-as-Engram recall scale with the number of
inserted facts? We probe by per-fact \emph{independent} OPT
insertion (the per-user override deployment mode) on the
first $n$ facts of an XL corpus of 1\,000 USER + 1\,000 ORG
templated facts. ICL@1000 is the in-context ceiling at the
hardest end. Figure~\ref{fig:factscale} shows the full curves.

\textbf{Per-fact recall is approximately flat in $n$ up to 1000
facts.} The best d8 and d12@1280 cells hold $\geq 0.98$ top-1
from $n{=}100$ to $n{=}1000$. The crucial caveat is that this is
\emph{independent} per-fact OPT: each fact gets its own private
hash rows, with no within-table interference between facts---%
the per-user override deployment mode of
Section~\ref{sec:method}. \emph{In that setting, we observe no
recall ceiling up to 1{,}000 facts on a 625\,M-parameter base
LM.} The density ceiling appears under Joint OPT into a shared
table (Section~\ref{sec:density}): 68\% $\to$ 35\% top-1 from
100 to 1{,}000 facts at d12@768.

\subsection{Dense-size scaling at optimal config}
\label{sec:dense-scaling}

Pulling the four trained Mini-Engrams into one row each
(Figure~\ref{fig:dense-scaling}):

\begin{figure}[t]
  \centering
  \includegraphics[width=\linewidth]{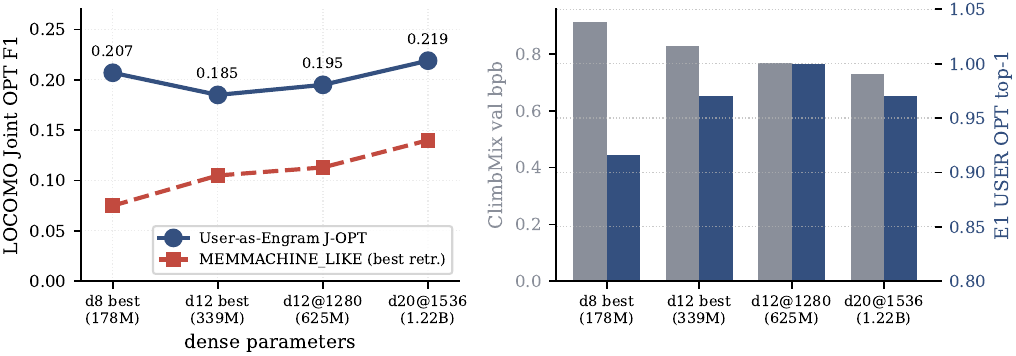}
  \caption{Dense-size scaling at the ablation-optimal recipe.
  Left: LOCOMO Joint OPT first-token token-F1 climbs from 0.134
  (d8 v2) to 0.233 (d20@1536); MEMMACHINE\_LIKE retrieval baseline
  climbs more slowly because its quality is bounded by
  sentence-encoder retrieval, not the LM. Right: val\_bpb (grey)
  drops monotonically with dense+tokens, while E1 USER OPT
  single-fact recall (blue) saturates at 1.00 from d12@1280 onward
  (small dip at d20 at iso-12 t/p).}
  \label{fig:dense-scaling}
\end{figure}


\paragraph{Three observations.}
(i) \textbf{Val bpb scales smoothly with dense $\times$ tokens}:
0.91 $\to$ 0.83 $\to$ 0.77 $\to$ 0.73.
(ii) \textbf{Insertion-OPT and E1 saturate} from d12@768 onward.
The 16-fact ins-OPT probe reaches 1.00 at every dense size with
${\geq}\,1.32$\,B tokens; E1 USER/ORG OPT reaches 1.00 at
d12@1280@3.34\,B, dipping slightly to 0.97 at
d20@1536@7.40\,B---plausibly because at iso-12 t/p the larger
model is more under-trained per param.
(iii) \textbf{LOCOMO Joint OPT first-token F1 scales
monotonically with dense size} from 0.134 (178\,M) to 0.233
(1.22\,B), and the multi-token token-F1 variant climbs from
0.159 to 0.310. Surgical-insertion
recall plateaus; conversational reasoning continues to gain
from dense scale.

\paragraph{Reading the gap.} Surgical insertion is a
\emph{lookup-then-cast} operation: the gate fires on the trigger
N-gram, the inserted row biases the gold next-token, and we
check rank 0. Once the base LM is fluent enough to make the gate
selective (${\geq}$ d12@768), recall saturates. LOCOMO, by
contrast, requires the LM to generate \emph{free-form} 16-token
answers conditioned on the Engram-anchored prefix; that
generation quality continues to improve with dense capacity.

\subsection{LoRA rank ablation at 100 facts}
\label{sec:lora-rank}

For completeness we also ablate the multi-fact LoRA rank at fixed
training budget (2000 steps, lr 5e-4, all $Q/K/V$ projections):


Every rank ${\geq}\,8$ saturates at the recall ceiling within a
2000-step budget. The interesting variable becomes
\emph{storage}, which scales linearly with rank. The takeaway:
at 100 facts/user, all rank-${\geq}\,8$ LoRAs hit the recall
ceiling, so the LoRA-vs-Engram trade-off collapses to
recall-vs-storage. \textbf{Engram Joint OPT is 20$\times$
smaller than the smallest LoRA at the ceiling (88\,KB vs.\
1.8\,MB rank-8), in exchange for a $\sim$35-pt top-1 gap
(65\% vs.\ 100\%; top-5 gap is 2\,pt: 98\% vs.\ 100\%).}
That gap closes by $n{=}1{,}000$, where LoRA rank-64 also slips
to 38\% top-1 (Section~\ref{sec:density}).

\subsection{Inserted-fact recall in free continuation}

Single-token recall measures whether the gold token is the
top-1 \emph{at the trigger position}. In conversational use the
model continues for many tokens; we test whether the inserted fact
surfaces in the first 8 generated tokens (greedy decoding) on
30 facts after Joint OPT (1500 steps).


\textbf{Dense scaling sharpens generation behavior.} At the
smallest scale the gold token is the immediate top-1 only 23\%
of the time but
\emph{anywhere in the next 8 tokens} 53\% of the time---the
model ``approaches'' the fact across a few tokens. From
d12@1280 onward, the gold either lands at position 0 or never
appears in the window: the bigger model commits earlier rather
than drifting. We caution that $n{=}30$ is small (Wilson 95\%
CI on 27/30 is $[74\%, 97\%]$; on 18/30 is $[42\%, 75\%]$), and
the result is consistent with the bigger model simply missing
the fact when its first guess is wrong. Either way, the 90\%
first-token rate at d20@1536 is the practical recall in
conversational use.

\subsection{Rank ablation: r=16 is the sweet spot}

\begin{figure}[t]
  \centering
  \includegraphics[width=0.5\linewidth]{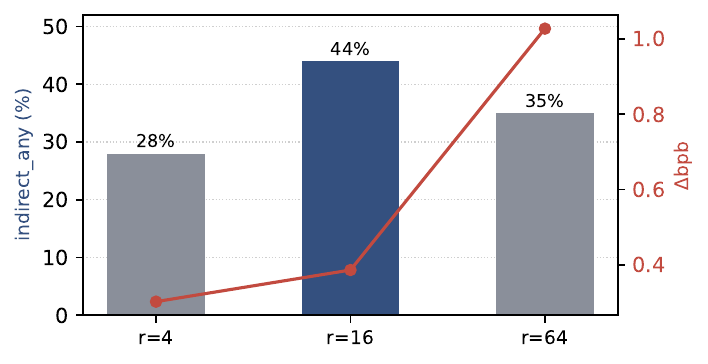}
  \caption{Shared-LoRA rank ablation (the layered design, $n{=}20$).
  The layered design
  preserves 100\% direct recall at every rank; indirect\_any
  peaks at $r{=}16$ (44\%) while contamination grows with rank.
  The $r{=}64$ surprise: $4\times$ the capacity gives
  \emph{worse} reasoning (35\%) at $2.7\times$ the contamination,
  plausibly because capacity exceeds what the 510-sample shared
  corpus can constrain.}
  \label{fig:shared-rank}
\end{figure}

We sweep the shared LoRA rank $r \in \{4, 16, 64\}$ on the full
20-user split (Figure~\ref{fig:shared-rank}) and observe a
clear, non-monotonic curve. Smaller ranks under-fit the reasoning skill;
larger ranks both contaminate more (more parameters, more global
perturbation) \emph{and} learn the reasoning skill worse, presumably
because the LoRA's capacity exceeds what the 510-sample training
corpus can profitably constrain.

The r=64 row is the surprise: a 4$\times$ larger shared LoRA
gives \emph{worse} indirect performance (35\% vs.\ 44\% at
r=16) at 2.7$\times$ more contamination. We recommend r=16;
future work could revisit this once the shared training corpus
is scaled beyond 510 samples.

\begin{figure}[t]
  \centering
  \includegraphics[width=0.85\linewidth]{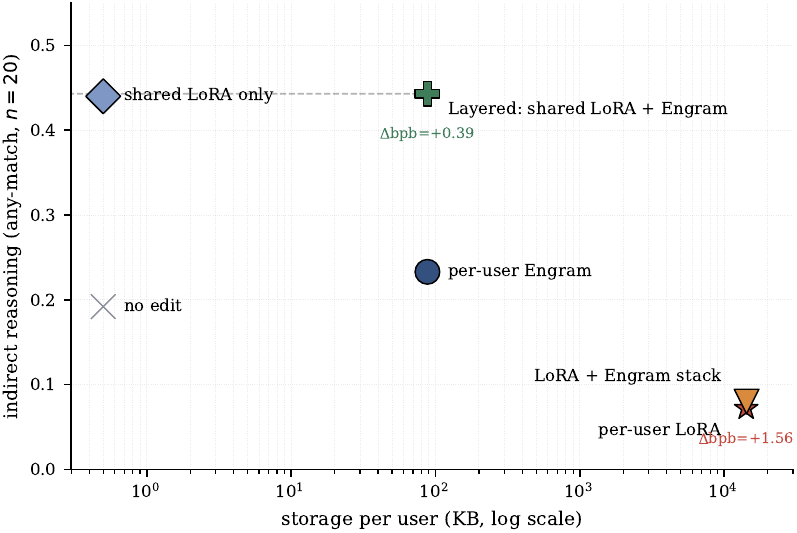}
  \caption{Cost-quality trade-off for the layered architecture
  (Section~\ref{sec:layered}). Same Mini-Engram-d20 base, $n{=}20$
  test users, six method combinations. \textbf{The layered design
  gives the
  best trade-off}: it matches the highest indirect-reasoning
  accuracy (44\% indirect\_any) at low storage (88\,KB/user) and low
  contamination ($\Delta$bpb $+0.39$). It beats per-user LoRA on
  every measure. The naive LoRA+Engram stack (the ``add Engram on
  top of LoRA'' baseline) is beaten too.}
  \label{fig:pareto-layered}
\end{figure}

\subsection{Cross-schema generalization}
\label{sec:layered-cross-schema}

\begin{figure}[t]
  \centering
  \includegraphics[width=0.62\linewidth]{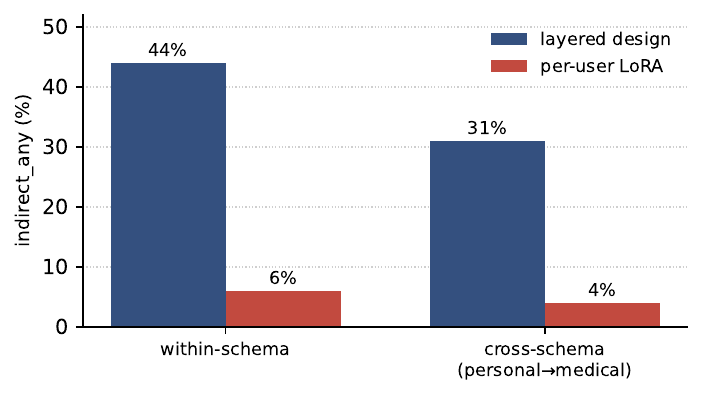}
  \caption{The layered claim survives a personal $\to$ medical
  schema shift (shared LoRA trained on personal-schema users,
  tested on medical-schema users). The layered design's
  indirect\_any decays from
  44\% to 31\% (a realistic 30\% relative drop) while per-user
  LoRA goes 6\% $\to$ 4\%, so the layered design's lead is preserved
  ($7.4\times \to 7.6\times$); locality is unchanged (the layered
  design's $\Delta$bpb equals the shared LoRA's, ${+}0.386$).}
  \label{fig:crossschema}
\end{figure}

The headline layered-design vs.\ per-user-LoRA numbers train the
shared LoRA on
held-out users in the same synthetic schema as the test users.
We test whether the reasoning skill transfers \emph{across}
schemas. We train the rank-16
shared LoRA on the original first-person personal-facts schema
(u020--u029; ``My name is\ldots'', ``I was born in\ldots'')
and evaluate the full six-condition head-to-head on a held-out
medical schema (m000--m019, third-person patient framing:
``The patient's name is\ldots'', ``MRN: \ldots''; 29 facts, 20
indirect probes per user; Figure~\ref{fig:crossschema}).

\paragraph{The layered claim survives cross-schema, with a
realistic decay.} The layered design's indirect\_any degrades from 44\%
(within-schema, Figure~\ref{fig:layered-conditions}) to 31\%
(cross-schema, Figure~\ref{fig:crossschema})---a 30\% relative drop. Over the
same gap, per-user LoRA's indirect\_any goes from 6\%
(within-schema) to 4\% (cross-schema), so the layered design's lead is
preserved ($7.4\times$ within-schema; $7.6\times$ cross-schema).
The architectural locality is guaranteed by design: the layered
design's $\Delta$bpb equals the shared LoRA's, $+0.386$, in both cases,
identical to four decimal places, because the per-user Engram
still fires only on its trigger N-grams.
\emph{Architectural decomposition transfers across schemas.}

\subsection{The layered design on a Q/A-format-adapted Engram base}
\label{sec:layered-sft}

The headline layered-design vs.\ per-user-LoRA numbers are on a base
LM (Mini-Engram-d20). Section~\ref{sec:negative}'s cross-base
table shows that per-user LoRA's \emph{user-visible} damage
shrinks dramatically on instruction-tuned bases (85\% of users
worse on a base LM, 0--20\% on four instruction-tuned bases).
Does the layered design still dominate per-user LoRA when the latter
itself becomes a stronger baseline?

We SFT Mini-Engram-d20 on $\sim$2000 single-token (Q, A) pairs
from the same fact templates the layered experiment uses (1500
steps; lr $1{\times}10^{-4}$; post-embedding params trained;
Engram tables frozen). The SFT is intentionally narrow---no
general-text corpus mixed in---so it tests Q/A format
adaptation rather than full chat-tuning. val\_bpb on raw
ClimbMix rises from 0.73 to 4.42, indicating that the model
has been pushed toward the Q/A distribution at the cost of
general-text LM ability. We then train a fresh rank-16 shared
LoRA on this SFT'd base
(\texttt{shared\_lora\_d20\_sft/r16}, 2000 steps on
u020--u029) and rerun the six-condition head-to-head; the
numbers below carry the conclusion.

\paragraph{The layered design still beats per-user LoRA on every
measure; the lead shrinks as the baseline
strengthens.} The qualitative picture survives Q/A adaptation:
the layered design matches per-user LoRA on direct (100\% vs.\ 100\%),
beats it on indirect
($1.6\times$ on top-1, $2.8\times$ on indirect\_any), and the
architectural locality is preserved (the layered design and the
shared LoRA alone differ in $\Delta$bpb by $1 \times 10^{-4}$, on
the same order as the $5 \times 10^{-5}$ original-base noise
floor). The cross-base prediction reproduces: per-user LoRA's
indirect\_top1 jumps from 1.3\% (untouched base) to 23.5\% on this
adapted base, matching Section~\ref{sec:negative}'s
``instruction-tuned bases absorb the LoRA perturbation''
pattern. The layered design's lead over per-user LoRA on
indirect\_any compresses from
$7.4\times$ (base LM) to $2.8\times$ (SFT'd), but it remains
the best method.

\paragraph{Caveats.}
The SFT here is narrow Q/A adaptation, not full chat-tuning:
mixing general-text in the SFT corpus or starting from a
true instruction-tuned base (the cross-base table's
Qwen/Llama/Mistral setup, but with an Engram-pretrained
backbone) is the natural follow-up. The aggressive SFT also
collapses the untouched base's and the per-user Engram's indirect to
floor (1\%), which inflates the layered design's relative lead over
the untouched base but not the layered-vs-LoRA contrast that
matters for deciding which method to use.

\subsection{Storage and cost sharing}

\begin{figure}[t]
  \centering
  \includegraphics[width=0.7\linewidth]{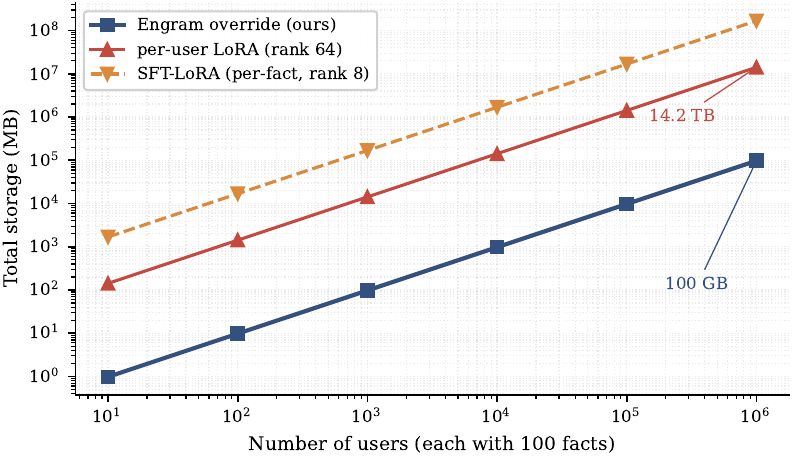}
  \caption{Per-user storage scales linearly with user count.
  Engram override is $\sim$161$\times$ smaller at 100 facts/user
  (88\,KB vs.\ 14.2\,MB) and the gap widens to $\sim$1700$\times$ at
  10 facts/user (LoRA's parameter count is fact-count-independent
  while Engram grows linearly). At 1\,M users $\times$ 100
  facts/user, Engram storage is 100\,GB vs.\ per-user LoRA's 14.2\,TB.}
  \label{fig:scaling}
\end{figure}

Per user, the layered design costs the Engram's 88\,KB; the
shared LoRA is one global 11.8\,MB (rank-16) file shared by all
users. For 1\,M users that is 100\,GB $+$
12\,MB $\approx$ 100\,GB, indistinguishable from Engram alone,
versus 14.2\,TB for per-user LoRA at the same recall ceiling.
Figure~\ref{fig:pareto-layered} plots all six method
combinations against storage and indirect-reasoning accuracy.

\end{document}